\documentclass[11pt]{article}

\usepackage[preprint]{acl}
\usepackage{times}
\usepackage{latexsym}
\usepackage[T1]{fontenc}
\usepackage[utf8]{inputenc}
\usepackage{microtype}
\usepackage{inconsolata}
\usepackage{graphicx}
\usepackage{booktabs}
\usepackage{amsmath}
\usepackage{amssymb}
\usepackage{array}
\usepackage{placeins}
\usepackage{float}
\usepackage{enumitem}

\graphicspath{{figures/}{figures/appendix/}}

\newcommand{\NA}{--}

\title{Controlled Memory Interference in Continual LLM Agents}
\author{
    \textbf{Ao Ding}\textsuperscript{1,*},
    \textbf{Hongzong Li}\textsuperscript{2,3,*},
    \textbf{Shiqin Tang}\textsuperscript{4},
    \textbf{Li Zhang}\textsuperscript{5},\\
    \textbf{Liang Chen}\textsuperscript{6},
    \textbf{Xuyang Chen}\textsuperscript{7},
    \textbf{Zi Liang}\textsuperscript{5,\textdagger}\\
    {\small \textsuperscript{1}China University of Geosciences (Beijing)\quad
    \textsuperscript{2}Northwestern Polytechnical University}\\[-2pt]
    {\small \textsuperscript{3}The Hong Kong University of Science and Technology\quad
    \textsuperscript{4}Chinese Academy of Sciences}\\[-2pt]
    {\small \textsuperscript{5}The Hong Kong Polytechnic University\quad
    \textsuperscript{6}École polytechnique fédérale de Lausanne}\\[-2pt]
    {\small \textsuperscript{7}National University of Singapore}\\[-2pt]
    {\scriptsize\ttfamily aoding2001@gamil.com; lihongzong@nwpu.edu.cn; lihongzong@ust.hk}\\[-2pt]
    {\scriptsize\ttfamily shiqin.tang@cair.cas.org.hk; zanly@mail.ustc.edu.cn; lchen@se.cuhk.hk}\\[-2pt]
    {\scriptsize\ttfamily xuyang.chen@nus.edu.sg; zi1415926.liang@connect.polyu.hk}
}

\begin{document}
\maketitle
\begingroup
\renewcommand{\thefootnote}{\fnsymbol{footnote}}
\footnotetext[1]{Equal contribution.}
\footnotetext[2]{Corresponding author.}
\endgroup

\begin{abstract}
Long-term memory enables AI agents to maintain continuity across
sessions, personalize behavior, and evolve through accumulated
experience. Yet memory evolution is not simply a process of storing
more information: new experiences may reinforce, revise, or interfere
with existing memory states. Existing systems mainly emphasize memory
construction and relevance-based retrieval, but several memories may
remain simultaneously relevant while differing in state, temporal
validity, or authority. We introduce \textbf{C}ontrolled \textbf{ M}emory \textbf{I}nterference (CMI), a controlled
diagnostic and data-generation framework for studying how agent memory
evolves under different memory relationships. Across controlled memory
evolution, benign accumulation has limited effects, whereas
relationship-specific interference sharply suppresses update
plasticity with little stability gain, either by blocking target-memory
exposure or by disrupting its downstream use. Lexical and Dense
retrieval exhibit distinct interference pathways, while poisoning is
more sensitive to update-authority cues than to recency alone. Beyond
diagnosis, CMI provides targeted examples for interference-aware
memory learning, improving the distinction between valid updates and
interference-inducing memories with no observed decrease in matched-unnoised point estimates. These findings show that memory evolution is shaped not only by
memory scale, but also by interactions among accumulated experiences.
More broadly, memory interference emerges as an important factor
for reliable continual agent memory systems.
\end{abstract}

\noindent\textbf{Code:}\\
{\scriptsize\raggedright\Urlmuskip=0mu plus 1mu\relax
\url{https://github.com/Dawn-Buendia/Controlled-Memory-Interference-in-Continual-LLM-Agents}\par}

\section{Introduction}
\label{sec:introduction}

Long-term memory is becoming a fundamental capability for large
language model (LLM)-based agents. Recent coding and interactive
agents increasingly support persistent,
tool-using, and long-running workflows, while memory-augmented designs
preserve experience across interactions
~\cite{park2023generative,packer2023memgpt,johnston2026shift}.
As agents operate across sessions and tasks, memory enables them to
preserve continuity, personalize behavior, and evolve through
accumulated experience. Memory is therefore more than an external
store: it provides the experiential basis through which an agent
remains consistent while continuing to adapt.

Existing agent-memory systems have made substantial progress in memory
construction, organization, updating, and access. Generative
Agents~\cite{park2023generative}, MemGPT~\cite{packer2023memgpt},
MemoryBank~\cite{zhong2023memorybank}, Mem0~\cite{mem0}, and
A-MEM~\cite{xu2026memories} provide representative approaches to
long-term memory management. Lexical, dense, and hybrid retrieval
methods further improve access to relevant historical information
~\cite{robertson2009probabilistic,karpukhin2020dense,
lewis2020retrieval,cormack2009reciprocal}. However, these approaches
primarily address whether relevant memories can be retained and
retrieved, rather than whether a retrieved memory remains appropriate
for the agent's current state.

\begin{figure*}[t]
    \centering
    \includegraphics[width=0.96\textwidth]{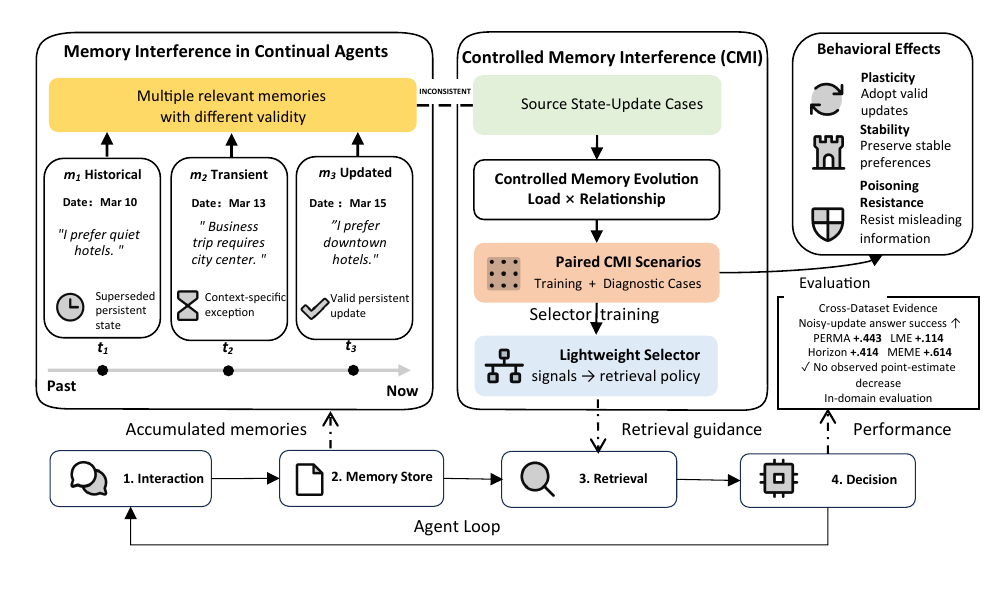}
    \caption{Overview of memory interference and CMI in the continual agent loop.
CMI controls memory load and relationships to construct paired
diagnostic and training scenarios, enabling behavioral evaluation and
lightweight retrieval adaptation. In-domain augmentation improves
noisy-update answer success across TravelPlanner and four external
sources.}
    \label{fig:overview}
\end{figure*}

As memory evolves, multiple experiences may remain simultaneously
relevant while differing in state, temporal validity, or authority.
An earlier preference may have been superseded by a valid update,
while repeated experiences may overemphasize a state that is no longer
appropriate. This problem is functionally analogous to proactive
interference and cue overload in human memory, where related prior
experiences obstruct newer information or make shared retrieval cues
less discriminative~\cite{underwood1957interference,
watkins1975cueoverload}. We refer to such failures in continual agents
as \textbf{memory interference}, without assuming that artificial and
human memory share the same internal mechanisms.
Figure~\ref{fig:overview} situates this challenge within the continual
agent loop and summarizes how CMI converts it into diagnostic and
training scenarios. Memory interference is broader than conventional retrieval failure.
Interference-inducing memories may prevent the target from entering
the retrieved context, or the target may be retrieved but fail to
guide the downstream decision. Failures can therefore arise across
the retrieval-to-decision pathway. Aggregate task accuracy, recall,
or ranking scores do not by themselves identify which memory
relationship caused an error or whether failure occurred before or
after target exposure.

To study this problem, we introduce
\textbf{Controlled Memory Interference (CMI)}, a controlled diagnostic
and data-generation framework for agent-memory evolution. CMI varies
memory scale and relationships among accumulated experiences while
preserving the query, target state, and expected decision. We
instantiate it as \textbf{CMI-Travel}, a controlled extension of the
Travel Planner tasks in MemoryArena~\cite{he2026memoryarena}. CMI-Travel
reconstructs memory trajectories with progressive load levels,
relationship-controlled interventions, paired behavioral probes, and
authority-sensitive poisoning conditions. It evaluates valid update
adoption (\emph{plasticity}), preservation of established states
(\emph{stability}), and resistance to misleading memories
(\emph{poisoning resistance}).

Our experiments uncover \textbf{selective plasticity suppression
during memory evolution}. Benign accumulation produces limited and
non-monotonic changes, whereas repeated history and same-slot
conflict substantially impair valid update adoption with little
corresponding stability gain. The phenomenon is therefore
relationship-specific and asymmetric rather than a general capacity
limitation or a symmetric stability--plasticity trade-off. A
retrieval-to-decision decomposition further shows that interference
may arise either before or after target exposure. Lexical Retrieval is
particularly sensitive to repeated surface cues and direct same-slot
conflict, whereas Dense retrieval is vulnerable to semantic
crowding among related memories. Poisoning is similarly role-sensitive:
persistent-update and authority cues are more influential than recency
alone.

Beyond diagnosis, CMI generates targeted examples for
\textbf{interference-aware memory learning}. Incorporating
CMI-generated data improves the distinction between valid updates and
interference-inducing memories with no observed decrease in matched-unnoised point estimates. Observable retrieval-state signals also support
adaptive lexical--dense access, improving target ordering and
downstream memory-state decisions. Memory interference is thus not
only measurable, but also an actionable target for improving continual
memory reliability.

Our contributions are summarized as follows:

\begin{enumerate}[label=\roman*.]
    \item We identify \textbf{\textit{selective plasticity suppression}} during memory
    evolution, a relationship- and architecture-dependent phenomenon
    that cannot be explained by memory scale alone.

    \item We introduce \textbf{CMI}, \textit{a portable controlled diagnostic} and
    \textit{data-generation framework}, and validate it across multiple
    datasets.

    \item We show that in-domain CMI augmentation improves target access,
    target rank, and noisy-update answer success, with no observed
    decrease in matched-unnoised point estimates.
\end{enumerate}

\section{Related Work}

\paragraph{Long-term memory and retrieval for agents.}
Agent-memory systems support long-running interaction through memory
storage, organization, reflection, and updating. Generative
Agents~\cite{park2023generative}, MemGPT~\cite{packer2023memgpt},
MemoryBank~\cite{zhong2023memorybank}, Mem0~\cite{mem0}, and
A-MEM~\cite{xu2026memories} represent major approaches to persistent
agent memory. Lexical, dense, and hybrid retrieval further improve
access to relevant history~\cite{robertson2009probabilistic,
karpukhin2020dense,lewis2020retrieval,cormack2009reciprocal}.
However, relevance-based access does not determine which memory is
appropriate when several retrieved experiences represent different
states or temporal roles.
Recent benchmarks evaluate long-horizon recall, state updates,
evolving preferences, and multi-entity dynamics
~\cite{wu2025longmemeval,he2026memoryarena,liu2026perma,
li2026horizonbench,jung2026meme}. Memory-poisoning studies further
show that untrusted stored examples can steer retrieval-augmented
agents~\cite{chen2024agentpoison}.

\paragraph{Interference in continual systems.}
Continual-learning research studies how new information disrupts
previously acquired knowledge, including the stability--plasticity
dilemma and methods such as EWC, GEM, and experience
replay~\cite{kirkpatrick2017overcoming,lopez2017gradient,
rolnick2019experience,parisi2019continual}. These works primarily
address interference during parameter learning or replay. We instead
study interactions among externally stored memories during retrieval
and downstream decision making.

\section{Methodology}
\label{sec:method}

\begin{figure*}[t]
    \centering
    \includegraphics[width=\textwidth]{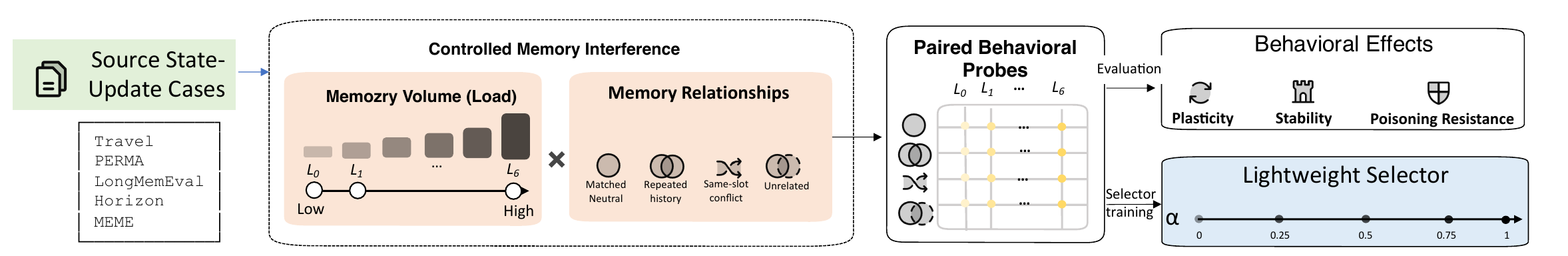}
    \caption{
CMI controls memory load and relationship composition to construct
paired behavioral probes and selector-training cases.
}
    \label{fig:framework}
\end{figure*}

We study memory interference in continual agents that accumulate
external memories over time. Several memories may be relevant to the
same decision while differing in state, temporal validity, or
authority. CMI isolates these factors through paired controlled
scenarios and uses the resulting cases for behavioral diagnosis and
interference-aware training, as summarized in
Figure~\ref{fig:framework}.

\subsection{Problem Formulation and Behavioral Dimensions}
\label{sec:problem}

Let $i$ index a source case. Each case contains a query $q_i$, an
expected decision $y_i$, and a memory collection
\[
\mathcal{M}_i=\{m_{i,1},\ldots,m_{i,N_i}\},
\]
where $m_{i,j}$ is the $j$-th memory and
$N_i=|\mathcal{M}_i|$ is the number of memories. Given a retrieval
depth $k$, a memory-access mechanism returns the ordered sequence
\[
R_k(q_i,\mathcal{M}_i)
=
\bigl(m_{i,(1)},\ldots,m_{i,(k)}\bigr),
\]
where $R_k$ denotes the access mechanism and $m_{i,(r)}$ is the
memory ranked at position $r$.

For evaluation, we partition the collection into three disjoint sets:
\[
\mathcal{M}_i
=
\mathcal{M}_i^{+}
\mathbin{\dot\cup}
\mathcal{M}_i^{c}
\mathbin{\dot\cup}
\mathcal{M}_i^{o},
\]
where $\dot\cup$ denotes disjoint union.
$\mathcal{M}_i^{+}$ contains the \emph{target memories} that should
influence the current decision, $\mathcal{M}_i^{c}$ contains
\emph{competing memories} that are relevant but inappropriate for the
current state, and $\mathcal{M}_i^{o}$ contains other controlled or
background memories, including neutral, unrelated, and task-compatible
same-state additions. Competing memories may encode superseded states,
transient exceptions, same-slot conflicts, or deliberately misleading
updates.

Memory interference occurs when relationships among these memories
alter which state influences the final decision. We evaluate three
behavioral dimensions. \textbf{Plasticity} measures whether the agent
adopts a valid persistent update instead of following a superseded
state. \textbf{Stability} measures whether it preserves an established
state when newer evidence is transient or context-specific.
\textbf{Poisoning resistance} measures whether it rejects an invalid
or misleading update. These dimensions represent distinct memory
roles and do not assume a symmetric stability--plasticity trade-off.

\subsection{Controlled Memory Interference}
\label{sec:controlled_evolution}

CMI applies to source cases in which an old state, a distinct current
state, and supporting historical evidence can be recovered. For each
eligible case $i$, we construct a base memory collection
$\mathcal{B}_i$ containing the target state and the context required
for the original decision. A controlled scenario is then defined as
\[
\mathcal{M}_i^{c,\ell}
=
\mathcal{B}_i\cup\Delta_i^{c,\ell},
\qquad
|\Delta_i^{c,\ell}|=n_{\ell},
\]
where $c\in\mathcal{C}$ indexes a memory-relationship condition,
$\ell\in\mathcal{L}$ indexes a memory-load level,
$\Delta_i^{c,\ell}$ is the set of controlled additions, and
$n_{\ell}$ is the number of additions at level $\ell$.
$\mathcal{C}$ and $\mathcal{L}$ denote the sets of evaluated
relationship conditions and load levels, respectively, and
$\mathcal{M}_i^{c,\ell}$ is the resulting CMI scenario.

All variants derived from the same source case preserve the underlying
task:
\[
q_i^{c,\ell}=q_i,\qquad
y_i^{c,\ell}=y_i,\qquad
\mathcal{M}_i^{+,c,\ell}=\mathcal{M}_i^{+},
\]
where $q_i^{c,\ell}$, $y_i^{c,\ell}$, and
$\mathcal{M}_i^{+,c,\ell}$ denote the query, expected decision, and
target-memory set in the transformed scenario. We additionally
preserve the target position and the source state semantics. Thus,
differences among paired variants can be attributed to the controlled
memory additions rather than changes in the underlying task.

CMI uses three complementary protocols. \textbf{Memory load} varies
accumulated history from L0 to L6. At fixed query and count,
\textbf{relationship composition} compares matched neutral additions,
unrelated memories, repeated history, and same-slot conflict
at counts 0, 2, 4, and 8. A separate \textbf{fixed-L3 authority probe}
frames an invalid memory as ordinary noise (P0), a recent transient
mention (P1), an explicit persistent update (P2), or an authoritative
correction (P3). Thus, P0--P3 are not additional conditions in the
four-way composition sweep. These protocols separate memory quantity,
scope, repetition, conflict, recency, and conveyed update authority.
The resulting \textbf{paired CMI scenarios} can serve as
diagnostic cases or as training examples. CMI does not assume that
arbitrary datasets can be transformed; the source case must provide a
recoverable state transition and sufficient evidence to identify the
superseded and current states.

\subsection{Diagnostic and Training Uses}
\label{sec:access_analysis}

\paragraph{Retrieval diagnostics.}
Let $r_i(m)$ denote the rank assigned to memory $m$ for case $i$. If
$m$ is absent from the returned top-$k$ sequence, we set
$r_i(m)=+\infty$. The best target and competing ranks are
\[
r_i^{+}
=
\min_{m\in\mathcal{M}_i^{+}}r_i(m),
\qquad
r_i^{c}
=
\min_{m\in\mathcal{M}_i^{c}}r_i(m).
\]
Let $\mathbb{I}[\cdot]$ denote the indicator function. We define
\[
\begin{aligned}
T_i
    &=\mathbf{I}[r_i^{+}\leq k],\\
C_i
    &=\mathbf{I}[r_i^{c}\leq k],\\
\mathrm{PO}_i
    &=\mathbf{I}\!\left[
      r_i^{+}\leq k
      \land
      \left(r_i^{c}>k\lor r_i^{+}<r_i^{c}\right)
      \right],\\
\mathrm{MRR}_i
    &=
    \begin{cases}
    1/r_i^{+}, & r_i^{+}<+\infty,\\
    0, & \text{otherwise},
    \end{cases}\\
\mathrm{CE}_i
    &=T_iC_i.
\end{aligned}
\]
Here, $T_i$ indicates target exposure, $C_i$ indicates
competing-memory exposure, $\mathrm{PO}_i$ indicates that the best
target outranks every retrieved competitor, $\mathrm{MRR}_i$ is the
reciprocal rank of the best target, and $\mathrm{CE}_i$ indicates
co-exposure. Dataset-level Target Recall@$k$, Preference Ordering,
MRR, and Co-exposure are obtained by averaging the corresponding
per-case quantities. MRR follows the standard reciprocal-rank
evaluation convention~\cite{voorhees2000trec8}; Preference Ordering
and Co-exposure are defined for this study.

\paragraph{Retrieval-to-decision decomposition.}
Let $A_i\in\{0,1\}$ denote answer success, with $A_i=1$ when the
agent output agrees with the expected decision $y_i$. Under the
diagnostic evaluator, target absence implies failure, so $A_i=0$
whenever $T_i=0$. If $A$ and $T$ denote the Bernoulli variables
obtained by uniformly sampling a case and taking its values $A_i$ and
$T_i$, then
\[
\Pr(A=1)
=
\Pr(T=1)\Pr(A=1\mid T=1).
\]
We additionally report
$\Pr(A=1\mid T=1,C=1)$, where $C$ is the Bernoulli variable induced
by $C_i$, to distinguish target-access failure from downstream
ambiguity after both target and competing memories enter the context.
These conditional quantities are descriptive diagnostics rather than
causal effects.

\paragraph{CMI-guided retrieval adaptation.}
We instantiate CMI-guided adaptation as an
\emph{Interference-Aware Adaptive Retrieval} (IAAR) selector.
We consider two complementary access families:
\textbf{Lexical Retrieval}, which uses token-level matching signals,
and \textbf{Dense Retrieval}, which uses semantic similarity in an
embedding space. Their concrete implementations are specified in
Section~\ref{sec:access_setup}.

For case $i$ and memory $m$, let
$\widehat{s}_{\mathrm{L}}(q_i,m)$ and
$\widehat{s}_{\mathrm{D}}(q_i,m)$ denote normalized Lexical and Dense
Retrieval scores. A lightweight selector $g$ maps an observable
retrieval-state feature vector $\mathbf{x}_i$ to a mixture weight:
\[
\alpha_i=g(\mathbf{x}_i),
\qquad
\alpha_i\in\{0,0.25,0.5,0.75,1\}.
\]
The combined score is
\[
s_i(m)
=
\alpha_i\widehat{s}_{\mathrm{L}}(q_i,m)
+
(1-\alpha_i)\widehat{s}_{\mathrm{D}}(q_i,m),
\]
where $s_i(m)$ is the final retrieval score,
$\alpha_i=1$ gives pure Lexical Retrieval, and $\alpha_i=0$ gives
pure Dense Retrieval. The feature vector $\mathbf{x}_i$ contains only
observable retrieval-state signals, such as score margin, retrieval
entropy, lexical overlap, semantic similarity, duplicate density, and
cross-retriever agreement. Target and competing-memory annotations are
not available to the deployed selector.

To test whether CMI scenarios provide useful supervision, we compare
clean-only training with in-domain CMI augmentation:
\[
\mathcal{D}_{\mathrm{aug}}
=
\mathcal{D}_{\mathrm{clean}}
\cup
\mathcal{D}_{\mathrm{CMI}},
\]
where $\mathcal{D}_{\mathrm{clean}}$ is the original selector-training
set, $\mathcal{D}_{\mathrm{CMI}}$ contains CMI-generated noisy
same-slot cases, and $\mathcal{D}_{\mathrm{aug}}$ is the augmented
training set. The resulting selector is evaluated on held-out cases
from the same source dataset. This protocol tests in-domain
actionability and does not assume zero-shot transfer of a frozen
selector across datasets.

\section{Experimental Setup}
\label{sec:experiments}

\subsection{Datasets and Protocols}
\label{sec:data_protocols}

We apply CMI to state-update cases derived from TravelPlanner in
MemoryArena~\cite{he2026memoryarena}, PERMA~\cite{liu2026perma},
LongMemEval~\cite{wu2025longmemeval},
HorizonBench~\cite{li2026horizonbench}, and the Tracking task from
MEME~\cite{jung2026meme}. TravelPlanner is used for fine-grained mechanism
analysis, including memory-scale growth, relationship-controlled
composition, retrieval-to-decision decomposition, and poisoning.
The remaining sources test whether CMI-generated same-slot
interference cases provide useful in-domain training supervision.

For each transformed case, CMI preserves the query, target state, and
expected decision while varying memory load or relationships among
accumulated memories. Target, competing-memory, and relationship
annotations are used only for construction, training-label generation,
and offline evaluation; retrieval and answer generation observe memory
text only.

The TravelPlanner mechanism study contains 50 base scenarios:
1,400 noise-growth instances across seven load levels, two
relationship-controlled composition protocols of 800 instances each,
1,000 fixed-L3 poisoning instances, and a held-out retrieval diagnostic
with 10 base scenarios and 320 observations per method. PERMA,
HorizonBench, and MEME Tracking each use 50 eligible bases;
LongMemEval retains 46 of 50 candidate updates after requiring a
distinct recoverable old state, confidence of at least $0.8$, and a
supporting historical evidence span.

PERMA, HorizonBench, and MEME Tracking use 30/10/10 train,
development, and test bases, yielding 630 clean and 210 CMI training
cases, 280 development cases, and 280 test cases. LongMemEval uses
27/9/10 bases, yielding 567 clean and 189 CMI training cases, 252
development cases, and 280 test cases. All splits are grouped by
\texttt{base\_id} with zero overlap. TravelPlanner follows the same
clean-only versus CMI-augmented comparison on its audited held-out
split.

\subsection{Memory Access and Selector Training}
\label{sec:access_setup}

We evaluate two primary access families.
\textbf{Lexical Retrieval} is instantiated with
BM25~\cite{robertson2009probabilistic} using
$k_1=1.5$ and $b=0.75$.
\textbf{Dense Retrieval} uses Qwen3-Embedding-8B
~\cite{zhang2025qwen3embedding} with its native
4096-dimensional output, L2 normalization, and cosine similarity.
We also evaluate fixed Lexical--Dense fusion with $\alpha=0.25$ and
RRF~\cite{cormack2009reciprocal} with rank constant $\kappa=60$.
All methods retrieve the top three
memories from the complete candidate collection.

The encoder-robustness analysis repeats the same protocol with
E5-Mistral-7B-Instruct~\cite{wang2024improving},
NV-Embed-v2~\cite{lee2024nvembed}, and
Linq-Embed-Mistral~\cite{choi2024linq}, using each model's native
4096-dimensional representation and official query/document encoding
convention.

The lightweight selector is a class-balanced random
forest~\cite{breiman2001random} with 300 trees, minimum leaf size 2,
and random seed 42. It predicts
$\alpha\in\{0,0.25,0.5,0.75,1\}$, where $\alpha=0$ denotes pure
Dense Retrieval and $\alpha=1$ denotes pure Lexical Retrieval.
Selector labels maximize a development utility combining behavior
success, target MRR, and security with weights $1$, $0.2$, and $0.1$,
respectively; ties within $0.01$ prefer the smaller lexical weight.

Clean-only and CMI-augmented selectors use the same model, observable
retrieval-state features, development protocol, and held-out test set.
They differ only in whether CMI noisy same-slot cases are included
during training. Training and evaluation are performed separately
within each source dataset; the protocol does not assume zero-shot
transfer of a frozen selector.

\subsection{Answer Evaluation and Statistics}
\label{sec:implementation}

TravelPlanner composition and mirrored-stability experiments use
DeepSeek-v4-pro with temperature 0, reasoning disabled, and a
512-token output limit. Its fixed-L3 poisoning study follows an
earlier frozen DeepSeek-v4-flash protocol and is interpreted only
within that protocol~\cite{deepseek2026v4}.

The answer-level augmentation study uses DeepSeek-v4-pro with
temperature 0, reasoning disabled, and a 128-token output limit.
TravelPlanner and the four external sources each contribute 280
held-out cases evaluated under clean-only and CMI-augmented training,
yielding $5\times280\times2=2{,}800$ generated answers. We observe no
API or parsing failures.

TravelPlanner outputs are scored using an audited structured
memory-state parser, while the four external sources use a
deterministic normalized value matcher. Answer success requires the
expected state to be selected without adopting a competing state;
wrong-memory adoption records the converse error. Full evaluator rules
are provided in the appendix.

We report paired 95\% percentile confidence intervals using cluster
bootstrap over \texttt{base\_id}, resampling all variants derived from
the same base together~\cite{efron1993bootstrap,cheng2013cluster}.
The answer-level study uses 5,000 bootstrap
replicates with random seed 42. Source filters, complete feature
definitions, prompts, preprocessing details, and protocol-specific
statistics are provided in the appendix.

\section{Results and Analysis}
\label{sec:results}

We organize the analysis around four questions: whether memory scale
alone explains degradation, which memory relationships determine the
failure regime, where interference arises along the
retrieval-to-decision pathway, and whether CMI-generated cases provide
useful training supervision. Behavioral and retrieval protocols are
reported separately because they evaluate different objectives.

\subsection{Memory Scale and Relationship-Specific Interference}
\label{sec:results-growth}
\label{sec:results-composition}

We first separate memory quantity from memory relationships. Under
TravelPlanner noise-only growth, average plasticity changes from
0.510 at L0 to 0.455 at L6, while stability changes from 0.200 to
0.290. The trajectories are non-monotonic: plasticity reaches its
lowest aggregate value around L4 and partially recovers at later
loads. Thus, increasing memory quantity affects access but does not
produce a consistent collapse. Full L0--L6 curves are provided in the
appendix.

We next hold memory count fixed while varying how added memories relate
to the current decision state. The TravelPlanner composition protocol
compares matched neutral, unrelated, repeated-history, and same-slot
conflict conditions at addition counts
$0$, $2$, $4$, and $8$. All variants derived from the same base case
share the query, target state, target position, expected decision, and
count-0 baseline.

\begin{figure}[t]
    \centering
    \includegraphics[width=\columnwidth]
    {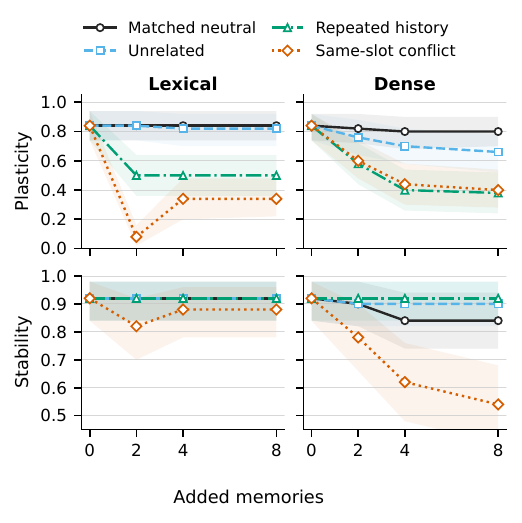}
    \caption{Relationship-controlled interference on TravelPlanner.}
    \label{fig:relationship-interference}
\end{figure}

Figure~\ref{fig:relationship-interference} reports plasticity and
mirrored stability across the four conditions.
Table~\ref{tab:travel_composition_effects} summarizes the count-8
outcomes relative to the shared count-0 baselines
($P_0=0.84$ and $S_0=0.92$).

\begin{table}[t]
\centering
\small
\setlength{\tabcolsep}{2.7pt}
\begin{tabular}{llrrrr}
\toprule
Access & Relationship & $P_8$ & $\Delta P$ & $S_8$ & $\Delta S$ \\
\midrule
Lexical
& Matched neutral
& .84 &  .00 & .92 &  .00 \\
& Unrelated
& .82 & $-.02$ & .92 &  .00 \\
& Repeated history
& .50 & $-.34$ & .92 &  .00 \\
& Same-slot conflict
& .34 & $-.50$ & .88 & $-.04$ \\
\midrule
Dense
& Matched neutral
& .80 & $-.04$ & .84 & $-.08$ \\
& Unrelated
& .66 & $-.18$ & .90 & $-.02$ \\
& Repeated history
& .38 & $-.46$ & .92 &  .00 \\
& Same-slot conflict
& .40 & $-.44$ & .54 & $-.38$ \\
\bottomrule
\end{tabular}
\caption{TravelPlanner count-8 behavior relative to shared count-0
baselines ($P_0=0.84$, $S_0=0.92$).}
\label{tab:travel_composition_effects}
\end{table}

Relationship, rather than count, determines the failure regime.
Lexical plasticity changes by only $-0.02$ under unrelated additions,
but by $-0.34$ under repeated history and $-0.50$ under same-slot
conflict. At count 8, the paired contrasts against unrelated additions
are $-0.32$ (95\% CI $[-0.46,-0.20]$) and $-0.48$
($[-0.62,-0.34]$). Dense Retrieval is more sensitive to semantic
crowding: unrelated additions reduce plasticity by $-0.18$, while
repeated history and same-slot conflict reduce it by $-0.46$ and $-0.44$.

The stability response is not a symmetric inverse of plasticity.
Repeated history leaves stability unchanged for both access families,
whereas Dense same-slot conflict reduces stability from 0.92 to 0.54.
Relationship-specific interference therefore ranges from selective
update suppression to joint degradation, rather than a universal
stability--plasticity trade-off.

\begin{figure}[t]
    \centering
    \includegraphics[width=\columnwidth]
    {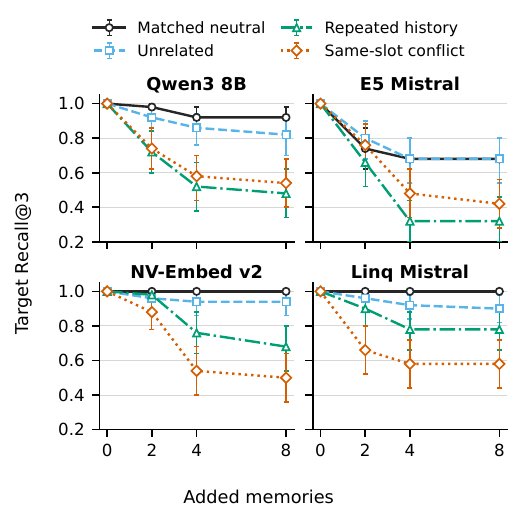}
    \caption{Encoder robustness under CMI. Target Recall@3 is shown
    as relationship-controlled memories are added. Error bars are
    paired base-cluster bootstrap 95\% CIs.}
    \label{fig:encoder-swap}
\end{figure}

Figure~\ref{fig:encoder-swap} shows that this relationship effect is
robust to the Dense encoder. Repeating
the fixed TravelPlanner retrieval protocol with Qwen3-Embedding-8B,
E5-Mistral-7B, NV-Embed-v2, and Linq-Embed-Mistral, all using native
unprojected 4096-dimensional representations, yields count-8
Target Recall@3 losses of 0.22--0.68 under repeated history and
0.42--0.58 under same-slot conflict. Thus, the qualitative
interference pattern is not specific to Qwen, although its magnitude
remains encoder dependent.

\subsection{Retrieval-to-Decision Pathways and Update Authority}
\label{sec:results-pathway}

Table~\ref{tab:travel_pathway} separates target-access failure from
failure after exposure in the count-8 TravelPlanner results.

\begin{table}[t]
\centering
\small
\setlength{\tabcolsep}{2.8pt}
\resizebox{\columnwidth}{!}{%
\begin{tabular}{llrrrr}
\toprule
Access & Relationship & Target R@3 & Comp.-only &
$P(A\mid T)$ & $P(A)$ \\
\midrule
Lexical & Matched neutral            & 1.00 & .00 & .840 & .840 \\
        & Unrelated          & 1.00 & .00 & .820 & .820 \\
        & Repeated history   & .66  & .00 & .758 & .500 \\
        & Same-slot conflict & .34  & .66 & 1.000 & .340 \\
\midrule
Dense   & Matched neutral            & .92  & .02 & .870 & .800 \\
        & Unrelated          & .82  & .18 & .805 & .660 \\
        & Repeated history   & .48  & .08 & .792 & .380 \\
        & Same-slot conflict & .54  & .46 & .741 & .400 \\
\bottomrule
\end{tabular}}
\caption{Count-8 retrieval-to-decision decomposition on TravelPlanner.}
\label{tab:travel_pathway}
\end{table}

Lexical same-slot conflict is primarily a retrieval bottleneck: Target Recall@3
falls to 0.34 and competing-only exposure rises to 0.66, but
$P(A\mid T)=1.00$. Repeated history combines retrieval loss
(Recall@3 of 0.66) with lower conditional adoption (0.758). Dense
Retrieval fails at both stages: repeated history lowers recall to
0.48, while same-slot conflict yields recall of 0.54 and conditional adoption of
0.741. Additional co-exposure diagnostics in the appendix show that
competitor presence alone is not a monotonic failure indicator. These
conditional quantities are descriptive rather than causal.

\paragraph{Update authority.}
At fixed L3, recency alone has negligible effect: poison acceptance
under P0/P1 is $.02/.02$ for Lexical Retrieval and $.00/.00$ for
Dense Retrieval. Framing the invalid memory as a persistent update
raises acceptance to $.23/.12$, and authoritative correction wording
further raises it to $.26/.14$. Thus, conveyed update role, rather
than recency alone, drives poisoning. Full protocol details are
reported in the appendix.

\subsection{Cross-Dataset Convergence}
\label{sec:results-external}

The external L0--L6 protocols progressively accumulate same-slot
historical states. Table~\ref{tab:external_growth} reports paired
displacement on PERMA, LongMemEval, HorizonBench, and MEME Tracking.

\begin{table}[t]
\centering
\small
\setlength{\tabcolsep}{3.4pt}
\begin{tabular}{lrrrr}
\toprule
& \multicolumn{2}{c}{Lexical} &
\multicolumn{2}{c}{Dense} \\
Source & $\Delta P$ & $\Delta S$ & $\Delta P$ & $\Delta S$ \\
\midrule
PERMA         & $-.720$ & $.000$  & $-.040$ & $.000$ \\
LongMemEval   & $-.826$ & $+.022$ & $-.043$ & $-.022$ \\
HorizonBench  & $-.940$ & $.000$  & $-.120$ & $.000$ \\
MEME Tracking & $-.860$ & $.000$  & $-.120$ & $.000$ \\
\bottomrule
\end{tabular}
\caption{L0--L6 displacement under same-slot historical growth.}
\label{tab:external_growth}
\end{table}

Lexical plasticity decreases by 0.720--0.940 while stability changes
by at most $+0.022$; all four plasticity intervals exclude zero.
Dense plasticity losses are smaller (0.040--0.120), with stability
nearly unchanged. Complete confidence intervals and noisy-update
results are provided in the appendix. These sources therefore show
convergent, not identical, selective update suppression under
accumulated superseded history.

\subsection{CMI-Guided Training Improves Retrieval and Final Answers}
\label{sec:results-actionability}

We compare clean-only and CMI-augmented selectors using identical
held-out cases, retrieval budgets, features, and model configurations.
Table~\ref{tab:cmi_actionability} reports noisy-update retrieval and
answer results.

\begin{table}[t]
\centering
\small
\setlength{\tabcolsep}{2.5pt}
\resizebox{\columnwidth}{!}{%
\begin{tabular}{lrrr|rrrr}
\toprule
& \multicolumn{3}{c|}{Retrieval change} &
\multicolumn{4}{c}{Answer success} \\
\cmidrule(lr){2-4}
\cmidrule(lr){5-8}
Source &
$\Delta$ R@3 &
$\Delta$ MRR &
$\Delta$ PO &
Clean-only &
CMI-aug. &
$\Delta$ &
95\% CI \\
\midrule
TravelPlanner
& $+.457$ & $+.252$ & $+.143$
& .386 & .686 & $+.300$ & [.129, .486] \\

PERMA
& $+.443$ & $+.168$ & $+.014$
& .386 & .829 & $+.443$ & [.243, .643] \\

LongMemEval
& $+.114$ & $+.041$ & $.000$
& .886 & 1.000 & $+.114$ & [.029, .214] \\

HorizonBench
& $+.414$ & $+.135$ & $+.014$
& .514 & .929 & $+.414$ & [.243, .600] \\

MEME Tracking
& $+.614$ & $+.426$ & $+.357$
& .314 & .929 & $+.614$ & [.429, .757] \\
\bottomrule
\end{tabular}}
\caption{Held-out noisy-update performance under clean-only and
in-domain CMI-augmented selector training.}
\label{tab:cmi_actionability}
\end{table}

CMI augmentation increases Target Recall@3 by 0.114--0.614 and MRR by
0.041--0.426 across all five sources; all paired intervals for these
metrics exclude zero. Preference Ordering improves less consistently,
and competing-memory recall remains high, so the selector improves
target inclusion and rank rather than fully reconciling competing
states.

Noisy-update answer success increases by 0.114--0.614, with every
paired interval excluding zero. Available clean, valid-update, and
stable-retention point estimates do not decrease. We do not claim
condition-wise non-inferiority because the LongMemEval
stable-retention paired interval is unavailable. The result supports
in-domain actionability, not complete reconciliation or zero-shot
cross-domain transfer.

\begin{figure}[t]
    \centering
    \includegraphics[width=\columnwidth]
    {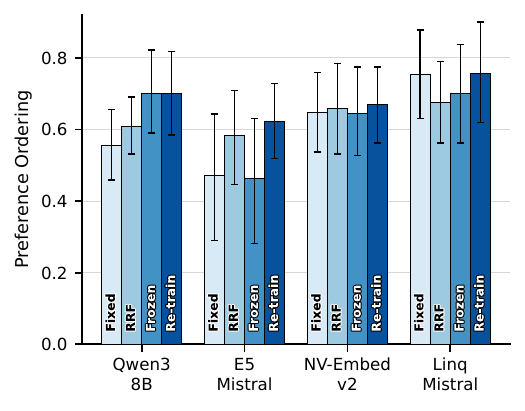}
    \caption{IAAR backend calibration. Preference Ordering compares
    fixed fusion, RRF, the frozen Qwen-trained selector, and a selector
    retrained on each backend. Error bars are base-cluster bootstrap
    95\% CIs.}
    \label{fig:iaar-calibration}
\end{figure}

Figure~\ref{fig:iaar-calibration} further bounds IAAR's scope through
a backend-calibration audit. Relative to
fixed fusion, the frozen Qwen-trained selector changes Preference
Ordering by $+0.144$ on Qwen but by $-0.009$, $-0.003$, and $-0.053$
on E5, NV, and Linq, respectively. Retraining the same selector and
feature schema per backend yields changes of $+0.150$, $+0.022$, and
$+0.003$ on E5, NV, and Linq. IAAR therefore exposes a calibration
opportunity rather than a universally transferable retrieval policy.

\section{Conclusion}
\label{sec:conclusion}

We introduced Controlled Memory Interference (CMI) to study how
relationships among accumulated memories affect continual agents.
TravelPlanner experiments separate memory quantity from repeated history,
same-slot conflict, and update authority, while retrieval-to-decision
analysis shows that interference can arise before or after target
exposure. Historical-state growth across PERMA, LongMemEval,
HorizonBench, and MEME further reveals architecture-dependent
plasticity suppression. Finally, in-domain CMI augmentation
consistently improves target access, target rank, and noisy-update
answer success across all five settings without an observed decrease
on matched unnoised point estimates. Preference ordering and
competitor exclusion remain open challenges for stronger
memory-state reconciliation. The interference pattern persists across
multiple native-4096 embedding backends, while adaptive retrieval
remains calibration-dependent rather than universally transferable.

\clearpage
\bibliography{references}

\clearpage
\appendix
\section*{Appendix}
\numberwithin{figure}{section}
\numberwithin{table}{section}
\numberwithin{equation}{section}
\section{Scope and Reproducibility Map}

The retained experimental scope comprises: (i) TravelPlanner-derived L0--L6
growth, relationship-controlled composition, mirrored stability, and fixed-L3
update-authority probes; (ii) CMI transformations of PERMA, LongMemEval,
HorizonBench, and the Tracking task from MEME; (iii) lexical, dense, fixed
fusion, and RRF access; (iv) clean-only versus
CMI-augmented IAAR training; and (v) native-4096 encoder robustness. Later
resolver, versioned-memory, representation-resolution, and router-optimization
branches are excluded.

Table~\ref{tab:checklist-map} maps the computational portions of the AAAI
reproducibility checklist to this appendix and the accompanying artifact
package. The paper makes no formal theoretical contribution, so the
conditional theory-only checklist items are intentionally left unanswered.

\begin{table*}[t]
\centering
\small
\setlength{\tabcolsep}{4pt}
\begin{tabular}{p{.08\textwidth}p{.13\textwidth}p{.70\textwidth}}
\toprule
Item & Status & Evidence supplied \\
\midrule
3.2--3.6 & yes/partial & Exact transformed CMI inputs, candidate mappings, source-specific protocol audits, eligibility rules, split protocol, upstream attribution, and redistribution notices. \\
4.1 & yes & Sections~\ref{sec:access}--\ref{sec:hyperparameters} list the development search space, selection criterion, and final settings. \\
4.2--4.3 & yes & The reviewer artifact's unpacked \path{code/}, \path{data/}, result records, and SHA-256 file manifest. \\
4.4--4.5 & partial & Author-written code is MIT licensed and executable scripts are included; transformed inputs retain source attribution and upstream terms, while model weights are not redistributed. \\
4.6 & partial & Seeds and deterministic rules are complete, while external API serving does not expose a reproducible sampling seed. \\
4.7 & yes & \texttt{seed\_map.md} and Section~\ref{sec:statistics}. \\
4.8 & yes & \texttt{environment.md} and Section~\ref{sec:environment}. \\
4.9--4.13 & yes & Metric definitions, run counts, clustered confidence intervals, paired comparisons, and final hyperparameters appear in Sections~\ref{sec:metrics}--\ref{sec:statistics}. \\
\bottomrule
\end{tabular}
\caption{Reproducibility-checklist map. Paths are relative to the uploaded
reviewer-artifact root unless stated otherwise.}
\label{tab:checklist-map}
\end{table*}

\section{Datasets and Memory Construction}
\label{sec:datasets}

CMI converts eligible state-update cases into paired continual-memory
trajectories. These are controlled transformations of source data rather than
evaluations under the sources' native leaderboard protocols. The transformation
keeps the base query, target state, and expected decision fixed while changing
memory load or relationships among memories. Gold target and competing-memory
annotations are used only for construction, selector-label generation, and
offline evaluation; retrieval and generation observe memory text only.

\subsection{Source Coverage and Eligibility}

Table~\ref{tab:data-coverage} gives the audited coverage. TravelPlanner-derived
cases come from MemoryArena. The external transformations use PERMA,
LongMemEval, HorizonBench, and the Tracking task from MEME. LongMemEval
retained 46 of 50 candidate knowledge
updates after requiring a recoverable old value distinct from the current
value, confidence at least 0.8, and a verbatim supporting span in the historical
session. The other external sources retained 50 eligible bases.

\begin{table*}[t]
\centering
\footnotesize
\setlength{\tabcolsep}{3pt}
\begin{tabular}{lrrrrl}
\toprule
Source & Bases & Growth rows & Composition rows & Constructed authority rows & Grouped split \\
\midrule
TravelPlanner & 50 & 1,400 & 800 P + 800 S & 400 & paired bases; held-out IAAR bases \\
PERMA & 50 & 1,400 & -- & 1,000 & 30/10/10 \\
LongMemEval & 46 & 1,288 & -- & 920 & 27/9/10 \\
HorizonBench & 50 & 1,400 & -- & 1,000 & 30/10/10 \\
MEME Tracking & 50 & 1,400 & -- & 1,000 & 30/10/10 \\
\bottomrule
\end{tabular}
\caption{Audited CMI data coverage. ``Growth rows'' count all four event types
over L0--L6. Travel composition contains separate plasticity and mirrored
stability datasets.}
\label{tab:data-coverage}
\end{table*}

For PERMA, HorizonBench, and MEME, the 30/10/10 base split yields 630
clean and 210 CMI training cases, 280 development cases, and 280 test cases.
LongMemEval's 27/9/10 split yields 567 clean and 189 CMI training cases, 252
development cases, and 280 test cases. All variants with the same
\texttt{base\_id} remain in one split; the supplied audits report zero base-ID
overlap.

Table~\ref{tab:data-coverage} distinguishes construction coverage from
answer-level evaluation coverage. All four external fixed-L3 datasets passed
the structural protocol audit and now have matched BM25/Dense answer-level
poisoning summaries under the same frozen generation and corrected-scoring
protocol.

\subsection{Transformation Audit and Representative States}

Every external source passed the same structural checks: the seven load
levels contain the expected number of memories; the four stress types remain
paired by base; target and forbidden states differ; memory identifiers are
unique; and no query contains a role annotation. Table~\ref{tab:construction-audit}
summarizes the machine-readable audit records. The L0 row contains the event's
minimum required memories (2--4); L1--L6 contain 6, 12, 24, 48, 90, and 135.

\begin{table}[t]
\centering
\small
\setlength{\tabcolsep}{3.5pt}
\begin{tabular}{lrrrr}
\toprule
Source & Rows & Bases & Failures & Pair failures \\
\midrule
PERMA & 1,400 & 50 & 0 & 0 \\
LongMemEval & 1,288 & 46 & 0 & 0 \\
HorizonBench & 1,400 & 50 & 0 & 0 \\
MEME Tracking & 1,400 & 50 & 0 & 0 \\
\bottomrule
\end{tabular}
\caption{External CMI protocol-audit results. Pairing failures are counted at
the base--stress group level.}
\label{tab:construction-audit}
\end{table}

Table~\ref{tab:state-examples} illustrates the source-dependent content
replacement. These examples are not evaluation outputs: they show how the
same CMI state-transition schema is instantiated in different domains.

\begin{table*}[t]
\centering
\small
\setlength{\tabcolsep}{4pt}
\begin{tabular}{>{\raggedright\arraybackslash}p{.16\textwidth}
                >{\raggedright\arraybackslash}p{.38\textwidth}
                >{\raggedright\arraybackslash}p{.17\textwidth}
                >{\raggedright\arraybackslash}p{.17\textwidth}}
\toprule
Source & Query & Historical state & Current state \\
\midrule
PERMA & User's favourite author? & Leo Tolstoy & Astrid Lindgren \\
LongMemEval & Wells Fargo mortgage pre-approval? & \$350,000 & \$400,000 \\
HorizonBench & Preferred response format? & bullet points & comparative synthesis tables \\
MEME Tracking & User's vehicle? & Therwyn Compact & Xylorim Scooter \\
\bottomrule
\end{tabular}
\caption{Representative audited old/current state pairs used by the external
CMI transformations.}
\label{tab:state-examples}
\end{table*}

\subsection{Record Structure and Retrieval View}

Each JSONL row stores a query, expected structured or normalized state,
session list, memory text, timestamp, session identifier, insertion order,
memory-role metadata, target IDs, and competing or forbidden IDs. Timestamps,
sessions, and role labels support construction and offline audits. Baseline
retrieval indexes only the \emph{memory text}; IAAR additionally uses observable
rank, score, text-overlap, embedding, count, and insertion-order statistics
defined in Section~\ref{sec:iaar}. No memory-ID suffix, event label, target ID,
or forbidden ID is provided to a deployed selector or generator.

Construction audits check duplicate memory IDs, query-to-memory answer leakage,
cross-load pairing, expected/forbidden identity, memory count, and role-label
leakage. External synthetic histories use SHA-256-derived deterministic content
selection, making the realized datasets independent of Python hash ordering.

\section{Controlled Interference Protocols}
\label{sec:protocols}

\subsection{L0--L6 Memory Growth}

The nominal load profile is L0--L6 with 3, 6, 12, 24, 48, 90, and 135
memories for the Travel growth summary. External event rows use the minimum
number of memories needed by the event at L0 (2--4), then the same L1--L6
targets of 6, 12, 24, 48, 90, and 135. Each base appears at every load. The
four event types are clean, valid update, stable retention, and valid update
with noise. Added history is held within the transformation protocol rather
than regenerated independently for each access mechanism.

\subsection{Relationship-Controlled Composition}

The paired composition study fixes query, target memory, gold state, target
position, and base scenario. It appends 0, 2, 4, or 8 memories under one of the
four relationships in Table~\ref{tab:composition-definitions}. Count zero is a
shared baseline; it is generated once per base and reused across the four
composition trajectories.

\begin{table}[H]
\centering
\small
\setlength{\tabcolsep}{4pt}
\begin{tabular}{p{.29\columnwidth}p{.62\columnwidth}}
\toprule
Relationship & Construction \\
\midrule
Matched neutral & Template- and length-matched memories that do not change the queried state. \\
Unrelated & Memories from non-overlapping entities or attributes, used to estimate capacity pressure. \\
Repeated history & Additional memories supporting the prior value in the same state trajectory without introducing a new value. \\
Same-slot conflict & Memories concerning the same entity--attribute slot but expressing a competing value. \\
\bottomrule
\end{tabular}
\caption{Paper terminology and construction of the four controlled memory
relationships.}
\label{tab:composition-definitions}
\end{table}

Plasticity asks whether a persistent valid update is adopted. The mirrored
stability protocol retains the same 50 bases and compositions but introduces a
one-occasion transient value and asks for the persistent/default state. Its
target is the stable value, not the transient exception. Retrieval is rerun for
the stability query; plasticity retrieval IDs are not reused.

\FloatBarrier
\subsection{Fixed-L3 Update Authority}

The fixed-L3 protocol isolates the authority conveyed by a misleading memory.
P0 supplies ordinary unrelated/control context, P1 gives the misleading value
recency without update language, P2 frames it as a persistent update, and P3
uses authoritative correction wording. Each condition contains 100 paired
cases for each reported access mechanism. This protocol is held separate from
L0--L6 growth and from relationship counts.

\FloatBarrier
\section{Metrics and Evaluators}
\label{sec:metrics}

\subsection{Retrieval Metrics}

Let $G_i$ be target memory IDs, $F_i$ competing or forbidden IDs, and $R_i^k$
the top-$k$ ranking for sample $i$, with $k=3$. We report
\begin{align}
T_i &= \mathbb{1}[G_i\cap R_i^k\neq\emptyset],\\
C_i &= \mathbb{1}[F_i\cap R_i^k\neq\emptyset],\\
\mathrm{MRR}_i &= \begin{cases}1/r_i^G,&r_i^G<\infty\\0,&\text{otherwise,}\end{cases}\\
\mathrm{CE}_i &= T_i C_i,
\end{align}
where $r_i^G$ is the best target rank. For selector evaluations,
preference ordering is
\begin{equation}
\mathrm{PO}_i=\mathbb{1}[T_i=1\ \wedge\ (C_i=0\ \vee\ r_i^G<r_i^F)].
\end{equation}
Target-only and competing-only rates are $T_i(1-C_i)$ and
$C_i(1-T_i)$. The score margin is the target score minus the strongest
forbidden score when both exist. Encoder-only robustness records the complete
ranking before truncation and uses target-versus-forbidden rank ordering; its
CSV retains this protocol-specific field separately.

\subsection{Behavior Metrics}

Plasticity is answer success on a valid persistent update. Stability is answer
success on the mirrored persistent-state query after a transient event. Noisy
plasticity is valid-update success with competing/noise memories present.
Poison acceptance is explicit adoption of the invalid state in the P0--P3
protocol; poisoning resistance is its complement only within that protocol.
Clean cases are not counted as stability.

For pathway diagnostics, $A_i$ denotes answer success and is analyzed as
$P(A\mid T)$ and $P(A\mid T,C)$. These conditional quantities localize failure
before versus after target exposure; they are descriptive diagnostics, not
causal effects.

\subsection{Generation Prompts and Parsing}

Travel composition uses the following invariant instruction, followed by the
slot type, selected memory context, query, and a slot-specific JSON schema:

\begin{quote}\small\ttfamily
You are a strict memory-state resolver.\\
Use only the stored memories below to answer the query.\\
If memories conflict, use the latest applicable stable update for the same
traveler/day/slot.\\
Do not explain. Do not copy long memory text. Return one valid compact JSON
object only.
\end{quote}

The mirrored stability prompt replaces the conflict sentence with: ``The query
asks for the persistent/default state. If a memory is temporary, one-time, or
for a single occasion, do not treat it as the default state.'' Food outputs use
\texttt{cuisine} and \texttt{selected\_option}; budget outputs use
\texttt{budget} and \texttt{selected\_option}; accommodation outputs use
\texttt{room\_type}, \texttt{joins\_traveler},
\texttt{rating\_preference}, \texttt{price\_preference}, and a summary of at
most 12 words. JSON extraction first parses the full response, then the first
enclosed JSON object. A missing or invalid object is a parse failure, never a
silent success.

The five-source answer-augmentation prompt is:

\begin{quote}\small\ttfamily
Answer the query using only the supplied memory context.\\
Resolve persistent updates and one-off exceptions carefully. Return JSON only:
\{"answer": "concise answer"\}.\\
Do not explain your reasoning and do not mention memory IDs.
\end{quote}

External values are case-folded, reduced to alphanumeric tokens, and compared
after whitespace normalization. A predicted value matches if it equals the
gold, contains the gold, or is a nonempty normalized substring of the gold with
length at least two. For list-valued LongMemEval states, every current item must
match and any superseded item is forbidden. Answer success requires a gold
match and no forbidden match. Incorrect-memory use requires explicit adoption
of a superseded value; it is not inferred as one minus answer success.

Travel uses a slot-aware structured parser and target-grounded scoring for the
composition mechanism analysis. The five-source augmentation study evaluates
the predicted state directly and does not require exposure of a particular
memory ID. This distinction is preserved in all result files.

\FloatBarrier
\section{Memory Access Mechanisms}
\label{sec:access}

\begin{table}[H]
\centering
\footnotesize
\setlength{\tabcolsep}{3pt}
\begin{tabular}{@{}p{.22\columnwidth}p{.70\columnwidth}@{}}
\toprule
Method & Final setting \\
\midrule
Lexical Retrieval & BM25 with $k_1=1.5$, $b=0.75$; case-folded tokenizer \texttt{[a-z0-9]+}; no stopword removal. \\
Dense Retrieval & Qwen3-Embedding-8B, native 4096 dimensions, attention-mask mean pooling over last hidden states, post-pooling L2 normalization, cosine similarity. \\
Fixed fusion & $s_i(m)=\alpha\tilde{s}_{i,B}(m)+(1-\alpha)\tilde{s}_{i,D}(m)$ with $\alpha=0.25$; $\alpha=1$ is pure lexical and $\alpha=0$ is pure dense. \\
RRF & One-based ranks and $s(m)=\sum_{r\in\{B,D\}}1/(60+\mathrm{rank}_r(m))$. \\
Shared budget & Top-3 for lexical, dense, fixed fusion, RRF, and IAAR; no reranker. \\
\bottomrule
\end{tabular}
\caption{Final access settings used by the retained experiments.}
\label{tab:access-settings}
\end{table}

The Qwen encoder maximum sequence length is 2048 in the external runner.
Equal retrieval scores are resolved deterministically by original insertion
order. Fixed fusion applies independent per-query min--max normalization over
the complete candidate collection:
\begin{equation}
\tilde{s}_r(m)=\frac{s_r(m)-\min_j s_r(j)}{\max_j s_r(j)-\min_j s_r(j)}.
\end{equation}
If the denominator is at most $10^{-12}$, every normalized score for that
retriever is set to zero. The development-only z-score candidate first computes
$(s-\mu)/\sigma$ and then min--max normalizes; a standard deviation or final
range at most $10^{-12}$ also produces all zeros.

The IAAR lexical implementation uses positive IDF
$\log(1+(N-df+0.5)/(df+0.5))$. This differs from the
\texttt{rank-bm25} baseline's negative-IDF behavior and is therefore recorded as
a separate implementation detail rather than assumed byte-identical.

\FloatBarrier
\section{IAAR Selector}
\label{sec:iaar}

IAAR is an exploratory selector, not a new memory architecture. It predicts
$\alpha\in\{0,0.25,0.5,0.75,1\}$ from inference-time observable retrieval
signals and applies the same normalized score fusion above. The deployed
feature schema contains the 18 features in Table~\ref{tab:iaar-features}.

\begin{table*}[t]
\centering
\small
\setlength{\tabcolsep}{4pt}
\begin{tabular}{p{.23\textwidth}p{.68\textwidth}}
\toprule
Features & Exact construction \\
\midrule
\path{bm25_top1_score}, \path{dense_top1_score} & Maximum raw score from each retriever. \\
\path{bm25_dense_top1_gap} & Maximum normalized lexical score minus maximum normalized dense score. \\
\path{topk_overlap} & Jaccard overlap of lexical and dense top-3 ID sets. \\
\path{retrieval_entropy} & Entropy of the equal-weight normalized lexical--dense candidate scores, defined below. \\
\path{score_margin} & Largest minus second-largest equal-weight combined score. \\
\path{candidate_semantic_similarity} & Mean raw dense score among dense top-3 candidates. \\
\path{lexical_overlap} & Mean query--memory token-set Jaccard over the complete candidate collection. \\
\path{duplicate_density} & Mean pairwise token-set Jaccard over the complete candidate collection. \\
\path{memory_age}, \path{memory_count}, \path{average_age} & Maximum insertion order, number of memories, and mean insertion order. \\
\path{retrieved_candidate_count} & Size of the lexical/dense top-3 union. \\
\path{topk_pairwise_disagreement} & Mean $(1-\cos(\mathbf{e}_a,\mathbf{e}_b))/2$ over pairs in the union. \\
\path{same_slot_candidate_count} & Number of union candidates matching an automatic query slot (food, budget, accommodation, transport, or activity; lexical fallback otherwise). \\
\path{automatic_contradiction_score} & Maximum same-slot disagreement multiplied by $0.5+0.5$ times the larger automatic temporal/update signal. \\
\path{duplicate_ratio} & Fraction of union pairs with lexical Jaccard at least 0.80. \\
\path{topk_similarity_range} & Maximum minus minimum raw dense score in the union. \\
\bottomrule
\end{tabular}
\caption{Frozen deployable IAAR feature schema. All candidate-set features use
the union of lexical and dense top-3 results unless stated otherwise.}
\label{tab:iaar-features}
\end{table*}

For retrieval entropy, let $u_j=.5\tilde{s}_{B,j}+.5\tilde{s}_{D,j}$ across
candidate IDs. The implementation shifts by the minimum,
$a_j=u_j-\min_\ell u_\ell$, and defines
\begin{equation}
p_j=\frac{a_j}{\sum_\ell a_\ell},\qquad
H=-\sum_{j:p_j>0}p_j\ln p_j.
\end{equation}
Natural logarithms are used. If any shifted value is non-finite or
$\sum_j a_j\le10^{-12}$, entropy is exactly zero.

The slot heuristic applies regular expressions for food, budget,
accommodation, transport, and activity. If no semantic slot matches, it uses
the first two normalized lexical tokens. The temporal signal counts four
pattern groups: current/update language, old-state language,
temporary/one-off language, and negation; its value is capped at one.

Inference forbids \texttt{event\_type}, \texttt{stress\_type}, composition
labels or counts, gold or forbidden IDs, target IDs, behavior outcomes, oracle
alpha, template IDs, and role-derived memory-ID features. The generator never
sees selector confidence, method names, retrieval scores, or these labels.

Oracle training labels maximize
\begin{equation}
\begin{aligned}
U_i(\alpha)={}&B_i(\alpha)+0.2\,\mathrm{MRR}_i(\alpha)\\
&+0.1\,\mathrm{Security}_i(\alpha),
\end{aligned}
\end{equation}
where $B_i$ is strict target-selected/forbidden-excluded retrieval success.
Utilities within 0.01 prefer the smaller lexical weight. The final selector is
a class-balanced random forest with 300 trees,
minimum leaf size 2, random state 42, and unrestricted parallel fitting. Bases
are split 60/20/20 with no cross-split variants. Clean-only and CMI-augmented
selectors share architecture, feature schema, development protocol, and test
set; only the addition of in-domain CMI noisy same-slot training cases changes.

\section{Hyperparameter Development}
\label{sec:hyperparameters}

Development compared normalization in \{min--max, z-score\}; selector families
were class-balanced logistic regression (maximum 3,000 iterations), the final
random forest, and a one-hidden-layer MLP (32 hidden units, maximum 500
iterations, early stopping). Development macro-F1 selected the classifier.
The fixed-fusion weight and normalization were selected on development data;
test data were not used for tuning. Backend-specific Qwen/E5/NV/Linq reruns
reuse the frozen random-forest configuration and perform no new hyperparameter
search.

Encoder robustness keeps native 4096-dimensional outputs for all models.
E5-Mistral-7B-Instruct and Linq-Embed-Mistral use last-token pooling and their
official query instruction; NV-Embed-v2 uses its native protocol.
All vectors are L2-normalized before cosine top-3 retrieval. Alternative
encoders use maximum length 512 and batch size 16; the primary Qwen runner uses
batch size 64 with out-of-memory backoff.

Travel composition and mirrored stability use \texttt{deepseek-v4-pro},
temperature zero, thinking disabled, and at most 512 output tokens. The
five-source answer experiment uses the same model and decoding controls with a
128-token limit. The frozen P0--P3 protocol uses
\texttt{deepseek-v4-flash}, temperature zero, API-default reasoning behavior,
and 512 output tokens.

\FloatBarrier
\section{Statistical Protocol}
\label{sec:statistics}

The paired unit is always the base scenario, not an individual transformed
row. For each comparison, all loads, compositions, access variants, or training
regimes derived from one \texttt{base\_id} are resampled together. We report
paired percentile 95\% cluster-bootstrap intervals. Table~\ref{tab:bootstrap-settings} records the actual
replicate count in the paper-facing files.

\begin{table}[H]
\centering
\small
\setlength{\tabcolsep}{3pt}
\begin{tabular}{lrrl}
\toprule
Result family & Reps. & Seed & Unit \\
\midrule
Travel growth & 5,000 & 42 & base ID \\
Travel composition point CIs & 1,000 & 20260721 & base ID \\
Travel paired contrasts & 5,000 & 42 & base ID \\
Pathway diagnostics & 2,000 & 42 & base ID \\
External growth & 5,000 & 42 & base ID \\
Selector retrieval & 5,000 & 42 & base ID \\
Answer augmentation & 5,000 & 42 & base ID \\
Encoder robustness & 1,000 & 42 & base ID \\
\bottomrule
\end{tabular}
\caption{Bootstrap configuration by result family.}
\label{tab:bootstrap-settings}
\end{table}

Initial Travel base selection uses seed 7. Composition base selection,
shuffling, shared-C0 ordering, and mirrored-stability sampling use seed
20260721. The fixed-L3 diagnostic bootstrap uses seed 20260708. IAAR splits,
classifiers, and the remaining paired analyses use seed 42. DeepSeek calls use
temperature zero, fixed prompts, fixed contexts, and fixed schemas, but the
retained API did not expose an explicit sampling seed. The reviewer artifact
therefore includes the realized per-sample generations and deterministic
evaluations; byte-identical new generation remains subject to provider-side
serving.

The five-source answer study evaluates 280 held-out cases under two training
regimes for each of five sources, giving $5\times280\times2=2{,}800$ generated
answers. It has no API or parsing failures. Reported means are point estimates
over held-out bases; bootstrap intervals are sampling uncertainty and are not
described as standard deviations across independent model runs.

\FloatBarrier
\section{Computing Environment}
\label{sec:environment}

Experiments ran on Ubuntu 22.04 (kernel 6.5.0-41) with two Intel Xeon Platinum
8352V processors (144 logical CPUs), 503 GiB RAM, and eight NVIDIA RTX 4090
GPUs with 24,564 MiB each. The NVIDIA driver was 555.58.02 and PyTorch reported
CUDA 12.8. Jobs used the Conda environment \texttt{reasoning} and available GPU
devices.

The captured versions were Python 3.10.19, PyTorch 2.10.0+cu128,
Transformers 5.12.1, scikit-learn 1.7.2, NumPy 2.2.5, pandas 2.2.3,
SciPy 1.15.3, Matplotlib 3.10.7, OpenAI client 2.15.0,
\texttt{rank-bm25} 0.2.2, \texttt{tiktoken} 0.12.0,
\texttt{accelerate} 1.14.0, \texttt{faiss-cpu} 1.13.2, FastAPI 0.128.0,
and Uvicorn 0.40.0. Model identifiers and provider-side revisions are recorded
as served at experiment time; no weights or API credentials are redistributed.

\FloatBarrier
\section{Additional Audited Results}
\label{sec:additional-results}

\subsection{Travel Memory Growth}

\begin{figure}[t]
\centering
\includegraphics[width=\columnwidth]{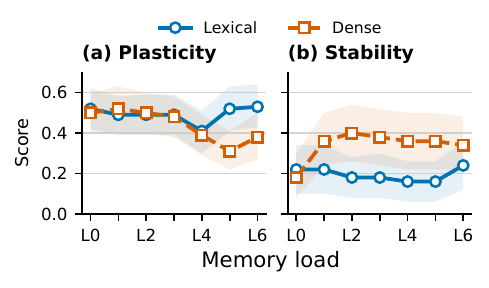}
\caption{Travel L0--L6 plasticity and mirrored stability. Error bars are 95\%
base-cluster bootstrap intervals.}
\label{fig:travel-growth}
\end{figure}

Figure~\ref{fig:travel-growth} reports both access families at every load with
base-cluster intervals. The trajectories are non-monotonic: scale changes the
operating point, but the severe relationship-specific failures in the next
subsection do not follow from count alone.

\begin{table*}[t]
\centering
\small
\setlength{\tabcolsep}{3.5pt}
\resizebox{\textwidth}{!}{%
\begin{tabular}{lrrrrrrr}
\toprule
& L0 & L1 & L2 & L3 & L4 & L5 & L6 \\
\midrule
Lexical P & .52 (.42,.62) & .49 (.39,.58) & .49 (.39,.59) & .49 (.39,.59) & .41 (.30,.51) & .52 (.41,.63) & .53 (.43,.64) \\
Lexical S & .22 (.10,.34) & .22 (.10,.34) & .18 (.08,.28) & .18 (.08,.30) & .16 (.06,.26) & .16 (.06,.26) & .24 (.12,.36) \\
Dense P & .50 (.40,.60) & .52 (.40,.63) & .50 (.40,.60) & .48 (.38,.58) & .39 (.29,.49) & .31 (.22,.41) & .38 (.27,.50) \\
Dense S & .18 (.08,.30) & .36 (.24,.50) & .40 (.26,.54) & .38 (.24,.52) & .36 (.22,.50) & .36 (.22,.50) & .34 (.22,.48) \\
\bottomrule
\end{tabular}}
\caption{Travel growth point estimates. Parentheses give 95\% base-cluster
intervals.}
\label{tab:travel-growth-full}
\end{table*}

\subsection{Relationship-Controlled Composition}

\begin{figure*}[t]
\centering
\includegraphics[width=.96\textwidth]{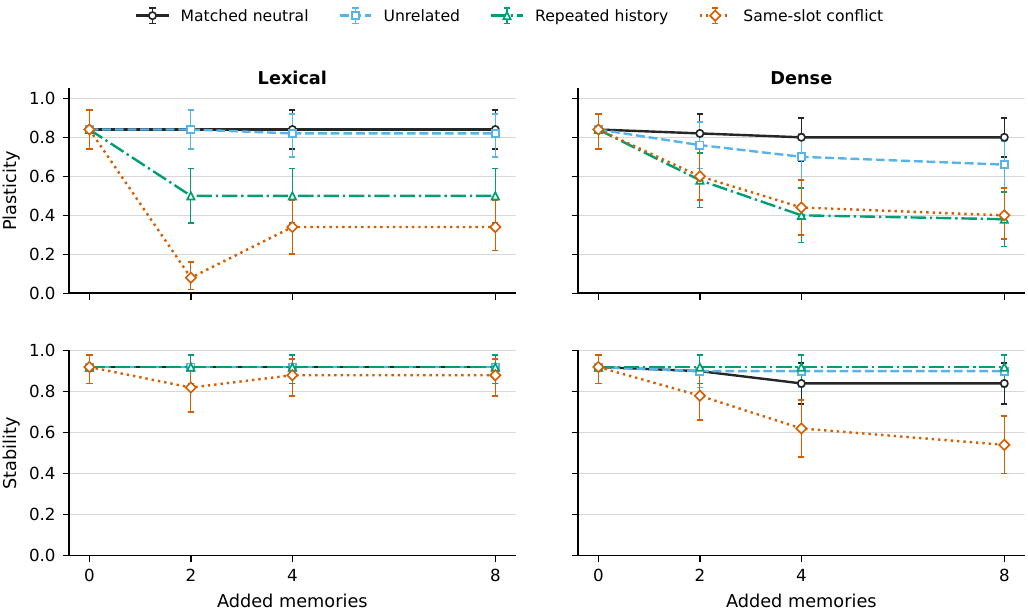}
\caption{Plasticity and mirrored stability for every relationship--count
ablation. Count zero is shared within each access family. Error bars are 95\%
base-cluster bootstrap intervals. Color, marker shape, and line style encode
the same relationship, preserving legibility under grayscale printing.}
\label{fig:composition-ps-curves}
\end{figure*}

Figure~\ref{fig:composition-ps-curves} plots every retained P/S ablation point
and its base-cluster interval. Table~\ref{tab:composition-full} gives the same
point estimates compactly. Table~\ref{tab:paired-deltas}
reports paired C8--C0 and C8 relationship contrasts; intervals are computed on
within-base differences rather than inferred from separate endpoint intervals.

\begin{table}[t]
\centering
\small
\setlength{\tabcolsep}{2pt}
\resizebox{\columnwidth}{!}{%
\begin{tabular}{llrrrrrrrr}
\toprule
Access & Relationship & P0 & P2 & P4 & P8 & S0 & S2 & S4 & S8 \\
\midrule
Lexical & Matched neutral & .84 & .84 & .84 & .84 & .92 & .92 & .92 & .92 \\
& Unrelated & .84 & .84 & .82 & .82 & .92 & .92 & .92 & .92 \\
& Repeated history & .84 & .50 & .50 & .50 & .92 & .92 & .92 & .92 \\
& Same-slot conflict & .84 & .08 & .34 & .34 & .92 & .82 & .88 & .88 \\
\midrule
Dense & Matched neutral & .84 & .82 & .80 & .80 & .92 & .90 & .84 & .84 \\
& Unrelated & .84 & .76 & .70 & .66 & .92 & .90 & .90 & .90 \\
& Repeated history & .84 & .58 & .40 & .38 & .92 & .92 & .92 & .92 \\
& Same-slot conflict & .84 & .60 & .44 & .40 & .92 & .78 & .62 & .54 \\
\bottomrule
\end{tabular}}
\caption{Plasticity (P) and stability (S) across memory relationships and
counts. The C0 point is shared within an access family.}
\label{tab:composition-full}
\end{table}

\begin{table*}[t]
\centering
\small
\setlength{\tabcolsep}{4pt}
\begin{tabular}{lllrrr}
\toprule
Access & Contrast & Metric & Estimate & CI low & CI high \\
\midrule
Lexical & C8--C0 repeated history & P & $-.34$ & $-.48$ & $-.22$ \\
Lexical & C8--C0 same-slot conflict & P & $-.50$ & $-.64$ & $-.36$ \\
Lexical & C8 repeated history minus unrelated & P & $-.32$ & $-.46$ & $-.20$ \\
Lexical & C8 same-slot conflict minus unrelated & P & $-.48$ & $-.62$ & $-.34$ \\
Dense & C8--C0 repeated history & P & $-.46$ & $-.60$ & $-.32$ \\
Dense & C8--C0 same-slot conflict & P & $-.44$ & $-.58$ & $-.30$ \\
Dense & C8 repeated history minus unrelated & P & $-.28$ & $-.46$ & $-.10$ \\
Dense & C8 same-slot conflict minus unrelated & P & $-.26$ & $-.40$ & $-.12$ \\
Dense & C8--C0 same-slot conflict & S & $-.38$ & $-.52$ & $-.24$ \\
\bottomrule
\end{tabular}
\caption{Key paired plasticity contrasts with 95\% base-cluster intervals
(5,000 resamples).}
\label{tab:paired-deltas}
\end{table*}

\subsection{Retrieval-to-Decision Pathway}

Table~\ref{tab:pathway-full} retains the complete C8 exposure partition. The
four exposure cells sum to one in every row. Conditional denominators are shown
because a high conditional success rate based on few exposed targets should
not be mistaken for high end-to-end success.

\begin{table*}[t]
\centering
\scriptsize
\setlength{\tabcolsep}{3pt}
\begin{tabular}{llrrrrrrrr}
\toprule
Access & Relationship & $P(T)$ & T-only & C-only & Co-exp. & Neither & $n_T$ & $P(A\mid T)$ & $P(A)$ \\
\midrule
Lexical & Matched neutral & 1.00 & .00 & .00 & 1.00 & .00 & 50 & .840 & .840 \\
& Unrelated & 1.00 & .00 & .00 & 1.00 & .00 & 50 & .820 & .820 \\
& Repeated history & .66 & .00 & .00 & .66 & .34 & 33 & .758 & .500 \\
& Same-slot conflict & .34 & .00 & .66 & .34 & .00 & 17 & 1.000 & .340 \\
\midrule
Dense & Matched neutral & .92 & .04 & .02 & .88 & .06 & 46 & .870 & .800 \\
& Unrelated & .82 & .00 & .18 & .82 & .00 & 41 & .805 & .660 \\
& Repeated history & .48 & .18 & .08 & .30 & .44 & 24 & .792 & .380 \\
& Same-slot conflict & .54 & .00 & .46 & .54 & .00 & 27 & .741 & .400 \\
\bottomrule
\end{tabular}
\caption{Count-8 retrieval-to-decision decomposition. $T$ is target exposure,
$C$ is competing exposure, and $A$ is answer success.}
\label{tab:pathway-full}
\end{table*}

\subsection{Update Authority}

\begin{figure}[t]
\centering
\includegraphics[width=\columnwidth]{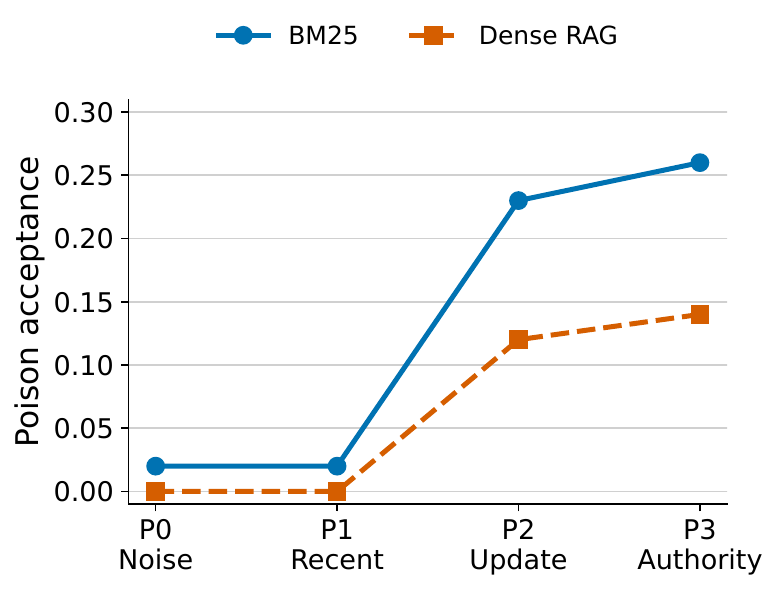}
\caption{Poison acceptance under fixed-L3 authority conditions. P0 is ordinary
control/noise, P1 adds recency only, P2 adds update wording, and P3 adds
authoritative correction wording.}
\label{fig:poison-authority}
\end{figure}

\begin{table}[H]
\centering
\small
\setlength{\tabcolsep}{5pt}
\begin{tabular}{lrrrr}
\toprule
Access & P0 & P1 & P2 & P3 \\
\midrule
Lexical & .02 & .02 & .23 & .26 \\
Dense & .00 & .00 & .12 & .14 \\
\bottomrule
\end{tabular}
\caption{Fixed-L3 poison-acceptance rates; 100 cases per access and condition.}
\label{tab:poison-full}
\end{table}

\subsection{Cross-Dataset Growth}

Table~\ref{tab:external-growth-full} reports L6--L0 answer-level displacement.
The paired intervals make the architecture contrast explicit: lexical access
shows large plasticity losses, while dense retrieval exhibits substantially
smaller changes on the same transformed bases.

\begin{figure*}[t]
\centering
\includegraphics[width=.90\textwidth]{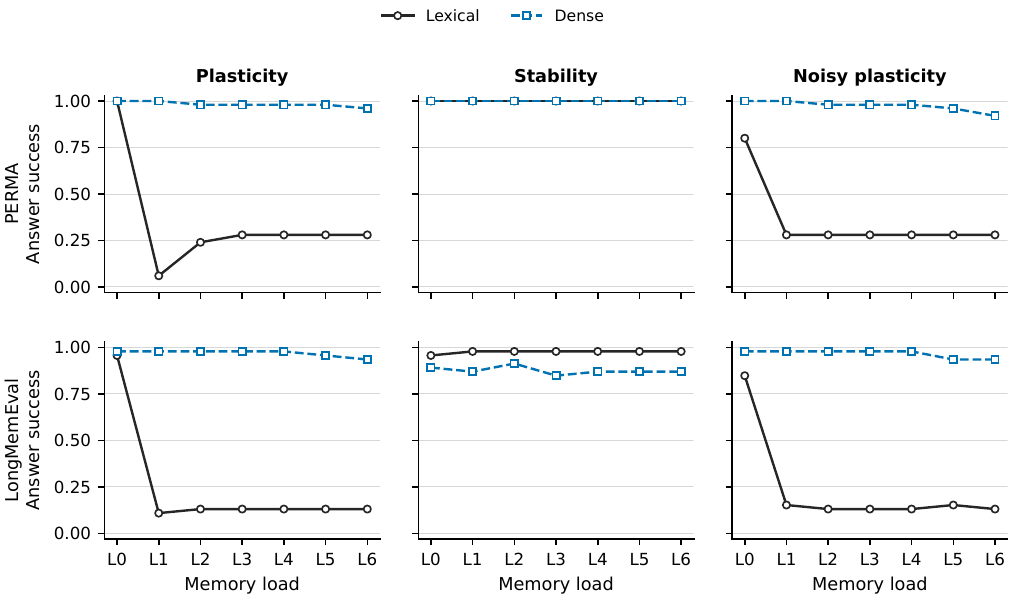}
\caption{Complete external L0--L6 trajectories retained for PERMA and
LongMemEval. Each panel uses the same paired bases across loads. Per-load
point estimates are shown without pseudo-replicate error bars; the paired
L6--L0 confidence intervals appear in Figure~\ref{fig:external-endpoint}.}
\label{fig:external-full-curves}
\end{figure*}

\begin{table*}[t]
\centering
\scriptsize
\setlength{\tabcolsep}{3.5pt}
\begin{tabular}{llrrr}
\toprule
Source & Access & $\Delta P$ & $\Delta S$ & $\Delta P_n$ \\
\midrule
PERMA & Lexical & $-.720$ $[-.840,-.600]$ & $.000$ $[.000,.000]$ & $-.520$ $[-.740,-.300]$ \\
& Dense & $-.040$ $[-.100,.000]$ & $.000$ $[.000,.000]$ & $-.080$ $[-.160,-.020]$ \\
LongMemEval & Lexical & $-.826$ $[-.935,-.696]$ & $+.022$ $[.000,.065]$ & $-.717$ $[-.891,-.522]$ \\
& Dense & $-.043$ $[-.109,.000]$ & $-.022$ $[-.065,.000]$ & $-.043$ $[-.109,.000]$ \\
HorizonBench & Lexical & $-.940$ $[-1.00,-.860]$ & $.000$ $[.000,.000]$ & $-.880$ $[-1.00,-.720]$ \\
& Dense & $-.120$ $[-.220,-.040]$ & $.000$ $[.000,.000]$ & $-.160$ $[-.260,-.060]$ \\
MEME Tracking & Lexical & $-.860$ $[-.940,-.760]$ & $.000$ $[.000,.000]$ & $-.860$ $[-.940,-.760]$ \\
& Dense & $-.120$ $[-.220,-.040]$ & $.000$ $[.000,.000]$ & $-.140$ $[-.240,-.060]$ \\
\bottomrule
\end{tabular}
\caption{Answer-level L6--L0 displacement with paired 95\% base-cluster
intervals. $P_n$ denotes valid-update success with noise.}
\label{tab:external-growth-full}
\end{table*}

\begin{figure*}[t]
\centering
\includegraphics[width=.74\textwidth]{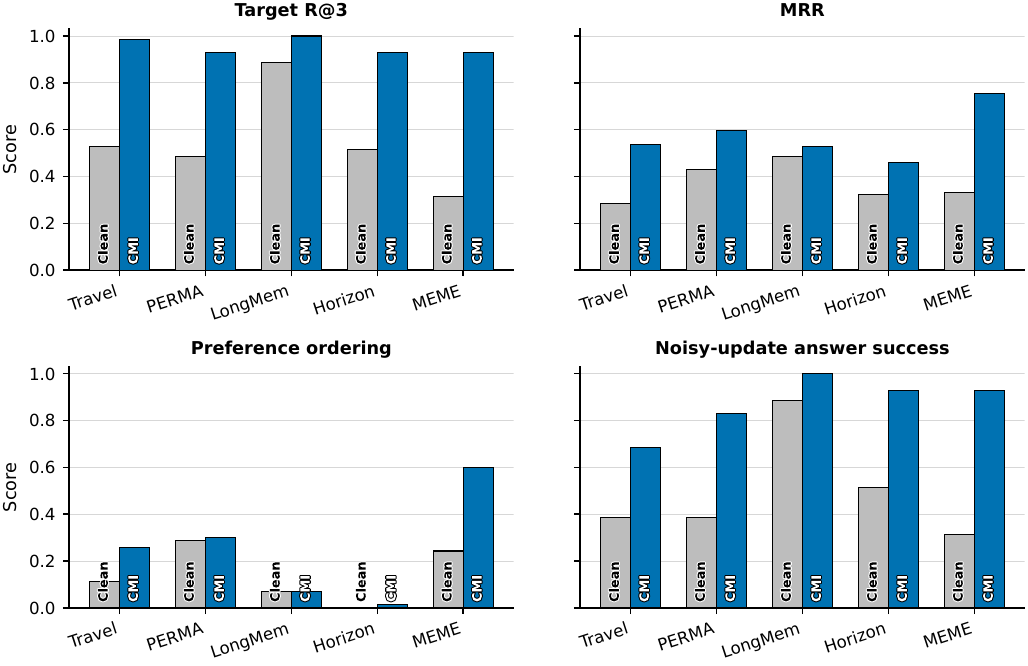}
\caption{Matched clean-only versus CMI-augmented IAAR results on five data
sources. The first three panels are retrieval-level noisy-update diagnostics;
the fourth is answer-level noisy-update success. Both training regimes use the
same held-out cases and selector architecture.}
\label{fig:iaar-all-sources}
\end{figure*}

IAAR is not inserted into this table because the retained selector experiment
uses a held-out noisy-update cohort rather than the same L0--L6 answer protocol.
Its matched retrieval and answer results are reported separately in
Figures~\ref{fig:iaar-all-sources} and~\ref{fig:iaar-vs-retrievers}.

\begin{figure*}[t]
\centering
\includegraphics[width=.90\textwidth]{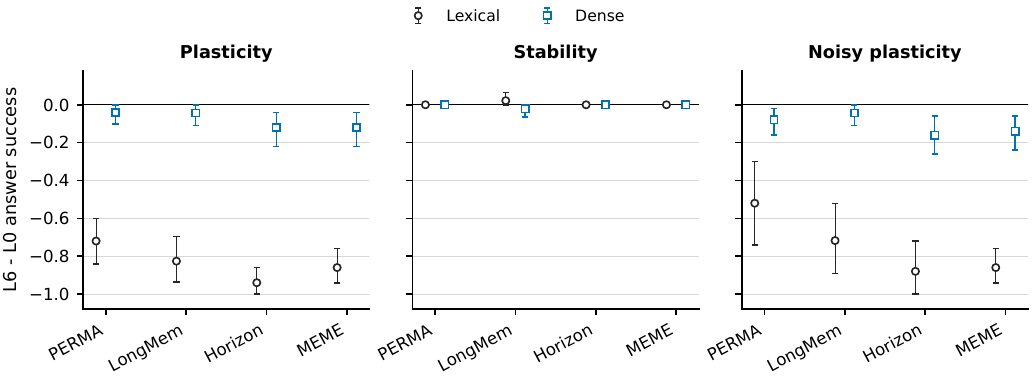}
\caption{L6--L0 answer-success displacement on all four external CMI
transformations. Error bars are paired 95\% base-cluster bootstrap intervals.
Full intermediate-load outputs were retained for PERMA and LongMemEval; the
HorizonBench and MEME archives retain audited endpoints and paired intervals.}
\label{fig:external-endpoint}
\end{figure*}

Complete endpoint estimates and paired intervals are retained in
\path{paper_results/chapter5/external_growth.csv}. In particular, all four
Lexical $\Delta P$ intervals exclude zero. These transformations diagnose
access under controlled historical growth; they do not replace each source's
native benchmark evaluation.

\subsection{CMI-Augmented Selector Retrieval}

\renewcommand{\dbltopfraction}{.94}
\setcounter{dbltopnumber}{2}

Table~\ref{tab:selector-full} reports the held-out noisy-update retrieval
diagnostic. CMI augmentation improves target access and MRR on all five sources,
while preference ordering does not improve uniformly because target and
competing memories can remain co-exposed.

\begin{table}[t]
\centering
\scriptsize
\setlength{\tabcolsep}{2pt}
\resizebox{\columnwidth}{!}{%
\begin{tabular}{llrrrrr}
\toprule
Source & Training & R@3 & MRR & PO & CE & Mean $\alpha$ \\
\midrule
TravelPlanner & Clean only & .529 & .283 & .114 & .529 & .296 \\
& CMI augmented & .986 & .536 & .257 & .986 & .411 \\
PERMA & Clean only & .486 & .430 & .286 & .486 & .768 \\
& CMI augmented & .929 & .598 & .300 & .929 & .293 \\
LongMemEval & Clean only & .886 & .486 & .071 & .886 & .689 \\
& CMI augmented & 1.000 & .526 & .071 & 1.000 & .532 \\
HorizonBench & Clean only & .514 & .324 & .000 & .514 & .782 \\
& CMI augmented & .929 & .459 & .014 & .929 & .171 \\
MEME Tracking & Clean only & .314 & .330 & .243 & .314 & .918 \\
& CMI augmented & .929 & .756 & .600 & .929 & .536 \\
\bottomrule
\end{tabular}}
\caption{Clean-only versus in-domain CMI-augmented IAAR retrieval on held-out
noisy updates. R@3 is target recall, PO is preference ordering, and CE is
co-exposure.}
\label{tab:selector-full}
\end{table}

\begin{table}[t]
\centering
\scriptsize
\setlength{\tabcolsep}{2pt}
\resizebox{\columnwidth}{!}{%
\begin{tabular}{lrrrrrr}
\toprule
Source & $\Delta$R@3 & 95\% CI & $\Delta$MRR & 95\% CI & $\Delta$PO & 95\% CI \\
\midrule
TravelPlanner & +.457 & [.314,.586] & +.252 & [.140,.383] & +.143 & [.000,.343] \\
PERMA & +.443 & [.243,.657] & +.168 & [.033,.287] & +.014 & [-.200,.229] \\
LongMemEval & +.114 & [.043,.200] & +.041 & [.011,.076] & .000 & [.000,.000] \\
HorizonBench & +.414 & [.243,.586] & +.135 & [.080,.200] & +.014 & [.000,.043] \\
MEME Tracking & +.614 & [.443,.757] & +.426 & [.268,.580] & +.357 & [.129,.586] \\
\bottomrule
\end{tabular}}
\caption{Paired CMI-augmentation deltas with 95\% base-cluster intervals.}
\label{tab:selector-deltas}
\end{table}

\begin{figure*}[t]
\centering
\includegraphics[width=.80\textwidth]{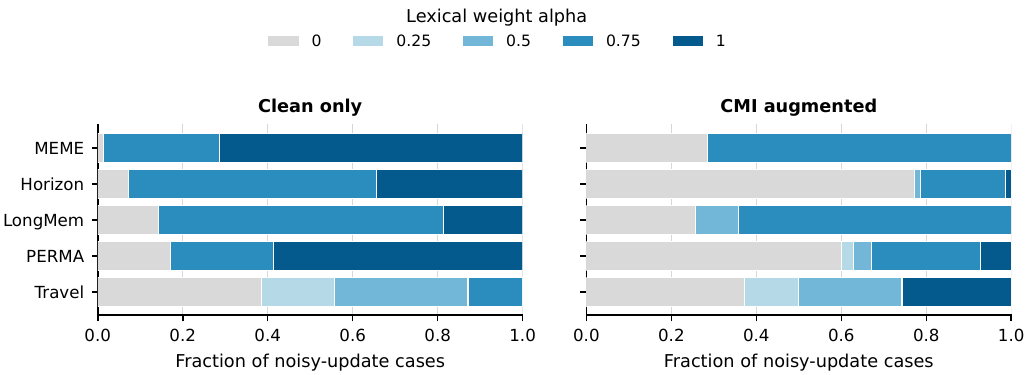}
\caption{IAAR lexical-weight distributions on held-out noisy-update cases.
Each bar contains 70 cases. Alpha 0 is pure dense retrieval and alpha 1 is pure
lexical retrieval. No event label is available to the selector at inference.}
\label{fig:iaar-alpha}
\end{figure*}

Figure~\ref{fig:iaar-alpha} shows that the retrieval gains do not arise from a
single constant alpha shift. CMI augmentation moves noisy-update cases toward
more dense or mixed access on PERMA and HorizonBench, while LongMemEval and
MEME retain substantial source-specific mass at alpha 0.75. This distribution
is descriptive; alpha itself is not assigned a semantic event label.

\begin{table}[t]
\centering
\small
\setlength{\tabcolsep}{2.5pt}
\resizebox{\columnwidth}{!}{%
\begin{tabular}{lrrrr}
\toprule
& \multicolumn{2}{c}{Clean only} & \multicolumn{2}{c}{CMI augmented} \\
Source & Accuracy & Macro F1 & Accuracy & Macro F1 \\
\midrule
TravelPlanner & .779 & .584 & .893 & .707 \\
PERMA & .786 & .432 & .839 & .581 \\
LongMemEval & .775 & .448 & .889 & .631 \\
HorizonBench & .746 & .519 & .921 & .714 \\
MEME Tracking & .789 & .716 & .914 & .906 \\
\bottomrule
\end{tabular}}
\caption{Held-out selector-label prediction under clean-only and in-domain
CMI-augmented training. These are all-event test metrics, not answer accuracy.}
\label{tab:router-training-allsource}
\end{table}

\subsection{IAAR Relative to Fixed Retrieval on Matched Cohorts}

\begin{figure*}[t]
\centering
\includegraphics[width=.74\textwidth]{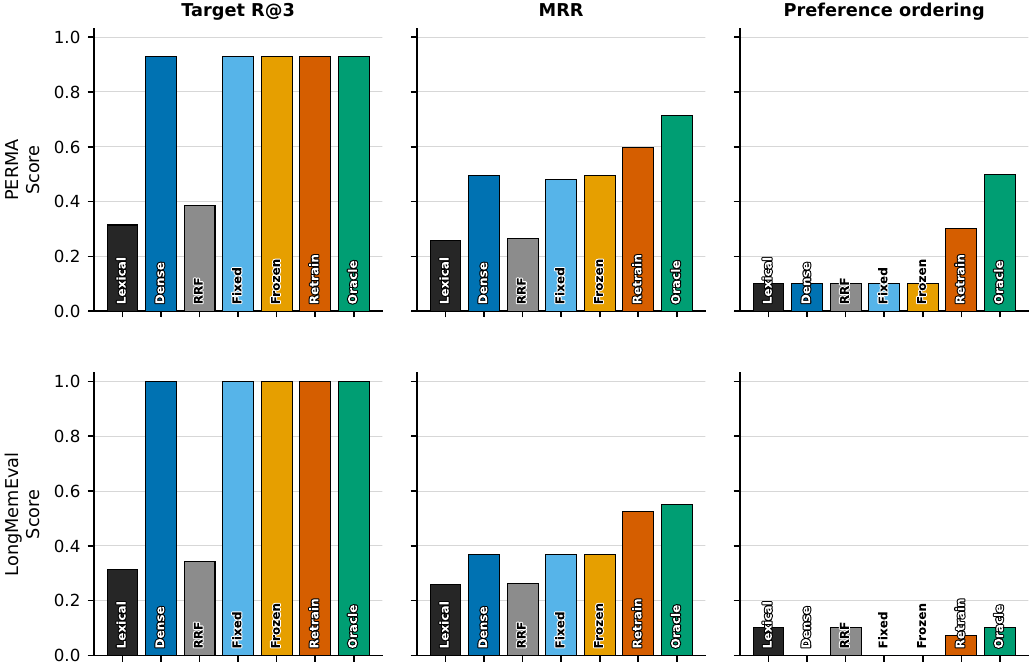}
\caption{Noisy-update retrieval comparison on matched PERMA and LongMemEval
test cohorts (70 cases per row). ``Frozen'' transfers the original selector;
``Retrain'' uses the source-specific fixed training protocol. The oracle is an
upper bound and is not deployable.}
\label{fig:iaar-vs-retrievers}
\end{figure*}

The aggregate paper-cutoff record contains a complete same-cohort comparison among BM25,
Dense RAG, RRF, fixed fusion, frozen IAAR, retrained IAAR, and an oracle for
PERMA and LongMemEval. Figure~\ref{fig:iaar-vs-retrievers} reports their
noisy-update retrieval metrics. We do not extend this figure to HorizonBench,
MEME, or Travel because the retained fixed-retriever summaries for those
sources are not on the exact held-out cohort used by the final selector
ablation. This scope restriction prevents an unmatched comparison.

Table~\ref{tab:router-audit} reports selector-label prediction separately from
end-to-end retrieval. Retraining greatly improves label prediction, but macro
F1 remains the relevant audit statistic because the five alpha classes are
imbalanced. These scores do not imply that the selected alpha dominates every
fixed retriever on every metric.

\begin{table}[H]
\centering
\footnotesize
\setlength{\tabcolsep}{2.5pt}
\begin{tabular}{llrrr}
\toprule
Source & Router & Acc. & Macro F1 & Weighted F1 \\
\midrule
PERMA & Frozen & .511 & .137 & .425 \\
& Retrained & .839 & .581 & .835 \\
LongMemEval & Frozen & .529 & .138 & .373 \\
& Retrained & .889 & .631 & .882 \\
\bottomrule
\end{tabular}
\caption{Frozen and source-retrained IAAR router metrics on held-out test
variants.}
\label{tab:router-audit}
\end{table}

\subsection{Answer-Level CMI Augmentation}

Table~\ref{tab:answer-full} reports all matched unnoised and noisy-update point
estimates. The noisy-update intervals are complete for all five sources. The
LongMemEval stable-retention interval and PERMA/LongMemEval absolute wrong-use
rates were not retained; they are left unavailable rather than inferred.

The Travel paired analysis in Table~\ref{tab:travel-answer-bootstrap} confirms
that the aggregate answer improvement is concentrated in noisy updates. The
unnoised interval includes zero and its point estimate is below 0.005, while
the noisy-update gain is 0.300. Incorrect-memory usage falls by the same paired
amount in this specific Travel evaluator; no complementarity assumption is
used for sources whose absolute wrong-use fields were not retained.

\subsection{External Fixed-L3 Poisoning Probe}

Table~\ref{tab:external-poison} reports all four external sources under the
matched answer-level fixed-L3 protocol. The non-monotonic source and access
differences reinforce that update framing, repetition, and the access mechanism
interact rather than inducing a universal severity order. Together with the
TravelPlanner P0--P3 result in Table~\ref{tab:poison-full}, this completes the
five-source fixed-L3 evidence retained in the appendix.

\begin{table*}[!t]
\centering
{\scriptsize
\setlength{\tabcolsep}{5pt}
\renewcommand{\arraystretch}{.88}
\begin{tabular}{llrrrr}
\toprule
Source & Condition & Clean & CMI aug. & Delta & 95\% CI \\
\midrule
TravelPlanner & Clean & 1.000 & 1.000 & .000 & [.000,.000] \\
& Stable retention & .700 & .700 & .000 & [.000,.000] \\
& Valid update & .871 & .886 & +.014 & [.000,.043] \\
& Noisy update & .386 & .686 & +.300 & [.129,.486] \\
PERMA & Clean & .900 & .900 & .000 & [.000,.000] \\
& Stable retention & .900 & .900 & .000 & [.000,.000] \\
& Valid update & .829 & .829 & .000 & [.000,.000] \\
& Noisy update & .386 & .829 & +.443 & [.243,.643] \\
LongMemEval & Clean & 1.000 & 1.000 & .000 & [.000,.000] \\
& Stable retention & .943 & .957 & +.014 & \NA \\
& Valid update & 1.000 & 1.000 & .000 & [.000,.000] \\
& Noisy update & .886 & 1.000 & +.114 & [.029,.214] \\
HorizonBench & Clean & 1.000 & 1.000 & .000 & [.000,.000] \\
& Stable retention & 1.000 & 1.000 & .000 & [.000,.000] \\
& Valid update & .986 & .986 & .000 & [.000,.000] \\
& Noisy update & .514 & .929 & +.414 & [.243,.600] \\
MEME Tracking & Clean & 1.000 & 1.000 & .000 & [.000,.000] \\
& Stable retention & 1.000 & 1.000 & .000 & [.000,.000] \\
& Valid update & .986 & .986 & .000 & [.000,.000] \\
& Noisy update & .314 & .929 & +.614 & [.429,.757] \\
\bottomrule
\end{tabular}}
\caption{Answer success under clean-only and in-domain CMI-augmented selector
training. Each row has 70 cases. ``CI'' is for the paired delta.}
\label{tab:answer-full}
\vspace{.3ex}

\begin{minipage}[t]{.42\textwidth}
\vspace{0pt}
\centering
\scriptsize
\setlength{\tabcolsep}{3pt}
\begin{tabular}{lrrr}
\toprule
Cohort and metric & Delta & CI low & CI high \\
\midrule
All: answer success & +.079 & +.036 & +.125 \\
All: incorrect use & $-.079$ & $-.122$ & $-.036$ \\
Unnoised: answer success & +.005 & .000 & +.014 \\
Noisy: answer success & +.300 & +.129 & +.486 \\
Noisy: incorrect use & $-.300$ & $-.486$ & $-.129$ \\
\bottomrule
\end{tabular}
\caption{Travel CMI-augmentation paired answer deltas (5,000 base-cluster
resamples).}
\label{tab:travel-answer-bootstrap}
\end{minipage}
\hfill
\begin{minipage}[t]{.53\textwidth}
\vspace{0pt}
\centering
\scriptsize
\setlength{\tabcolsep}{2pt}
\resizebox{\linewidth}{!}{%
\begin{tabular}{llrrr}
\toprule
Source & Access & Recency & Repeated history & Latest update \\
\midrule
PERMA & Lexical & .360 & .460 & .860 \\
& Dense & .060 & .465 & .810 \\
LongMemEval & Lexical & .440 & .516 & .489 \\
& Dense & .212 & .848 & .745 \\
HorizonBench & Lexical & .470 & .600 & .590 \\
& Dense & .115 & .920 & .775 \\
MEME Tracking & Lexical & .430 & .480 & .930 \\
& Dense & .055 & .370 & .765 \\
\bottomrule
\end{tabular}}
\caption{Poison-acceptance rates in the external fixed-L3 answer probe.
Recency adds one invalid recent state; repeated history adds three; latest
update gives one invalid memory explicit update status. Each access-condition
cell contains 184 LongMemEval cases or 200 cases for the other sources.}
\label{tab:external-poison}
\end{minipage}
\end{table*}

\subsection{Native-4096 Encoder Robustness}

\begin{figure*}[!t]
\centering
\includegraphics[width=.74\textwidth]{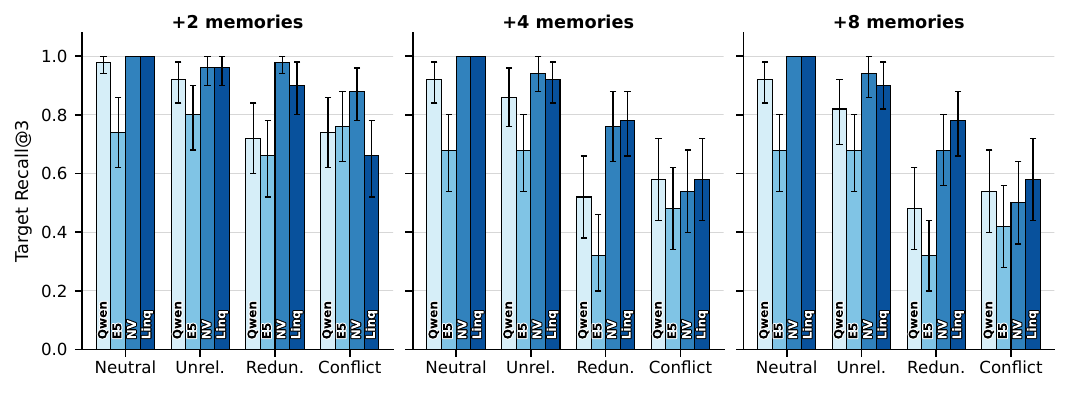}
\caption{Target Recall@3 at C2, C4, and C8 for four native-4096 embedding
backends. Labels are printed inside bars; color remains distinguishable in
grayscale through luminance.}
\label{fig:encoder-robustness}
\end{figure*}

Figure~\ref{fig:encoder-robustness} changes only the dense encoder while keeping
native 4096-dimensional output, cosine top-3 retrieval, query set, memory set,
and composition counts fixed. It is a robustness check rather than an encoder
leaderboard. Error bars use 1,000 base-cluster resamples.

Across encoders, matched neutral and unrelated additions are generally less
damaging than repeated history or same-slot conflict, although the magnitude
and non-monotonic details are backend dependent. The accompanying composition
summary and C8--C0 bootstrap CSV preserve the values shown in this figure;
fresh per-sample rankings are reconstructed by the retained encoder script.

\begin{table}[H]
\centering
\scriptsize
\setlength{\tabcolsep}{3pt}
\begin{tabular}{@{}llr@{}}
\toprule
Encoder & Relationship & $\Delta$R@3 [95\% CI] \\
\midrule
Qwen3-8B & Matched neutral & $-.08$ [$-.16,-.02$] \\
 & Unrelated & $-.18$ [$-.30,-.08$] \\
 & Repeated history & $-.52$ [$-.66,-.38$] \\
 & Same-slot conflict & $-.46$ [$-.60,-.32$] \\
\addlinespace
E5-Mistral & Matched neutral & $-.32$ [$-.44,-.20$] \\
 & Unrelated & $-.32$ [$-.44,-.20$] \\
 & Repeated history & $-.68$ [$-.80,-.56$] \\
 & Same-slot conflict & $-.58$ [$-.72,-.44$] \\
\addlinespace
NV-Embed-v2 & Matched neutral & $.00$ [$.00,.00$] \\
 & Unrelated & $-.06$ [$-.14,.00$] \\
 & Repeated history & $-.32$ [$-.46,-.18$] \\
 & Same-slot conflict & $-.50$ [$-.64,-.36$] \\
\addlinespace
Linq-Mistral & Matched neutral & $.00$ [$.00,.00$] \\
 & Unrelated & $-.10$ [$-.18,-.02$] \\
 & Repeated history & $-.22$ [$-.34,-.12$] \\
 & Same-slot conflict & $-.42$ [$-.56,-.28$] \\
\bottomrule
\end{tabular}
\caption{C8--C0 Target Recall@3 changes under native-4096 encoder replacement.
Brackets contain 95\% base-cluster bootstrap intervals.}
\label{tab:encoder-deltas}
\end{table}

The direction of the relationship effect is robust, but its magnitude is not
encoder invariant. In particular, Linq is comparatively robust to repeated
history whereas NV exhibits a larger same-slot-conflict loss. We therefore use
the encoder swap as evidence for a recurring relationship ordering, not for a
universal ranking among embedding models.

\FloatBarrier
\section{Threats to Validity and Evidence Boundaries}
\label{sec:threats}

\paragraph{Controlled transformations versus native benchmarks.}
CMI reuses states and queries from external sources but imposes a common
historical-growth protocol. The reported values diagnose memory interference;
they are not replacements for PERMA, LongMemEval, HorizonBench, or MEME native
leaderboard scores. Source-specific language can still affect lexical and
dense retrieval, which is why results are reported by source rather than only
as a pooled mean.

\paragraph{Coverage and statistical uncertainty.}
Each source contains 46--50 paired bases. Cluster bootstrap intervals quantify
uncertainty over those bases, not variation across independently trained
generators or embedding checkpoints. HorizonBench and MEME retain audited
L0/L6 endpoints and paired intervals but not every intermediate answer-level
summary in the final paper archive; no missing trajectory points are
interpolated. LongMemEval stable-retention augmentation lacks a retained paired
interval, so only its point estimate is reported.

\paragraph{Retrieval and generation scope.}
The primary dense access uses one 4096-dimensional Qwen encoder, with three
same-dimensional encoder replacements as a retrieval robustness check. The
answer studies use one hosted generator under deterministic decoding controls;
provider-side serving may still change. Retrieval-pathway conclusions are
therefore limited to the reported lexical, dense, fusion, RRF, and IAAR access
mechanisms.

\paragraph{Selector scope.}
IAAR is an exploratory selector trained on oracle-derived alpha labels, not a
general memory-state reasoner. Source retraining improves its label prediction
and noisy-update access, but preference ordering remains weak on several
sources because target and competitor can be co-exposed. The frozen-versus-
retrained full retriever comparison is available only for matched PERMA and
LongMemEval cohorts. We therefore claim evidence that CMI augmentation can
teach useful interference signals, not that IAAR universally dominates fixed
retrieval.

\paragraph{Annotation and redistribution.}
Gold target, historical, and forbidden roles are required to construct and
audit CMI but are hidden from deployed retrieval, selection, and generation.
LongMemEval old values include an API-assisted extraction followed by strict
evidence-span verification; 46 of 50 paper-cutoff candidates pass. The
confidential reviewer artifact includes the exact transformed inputs consumed
by the retained experiments, while upstream URLs are retained only for
attribution and provenance. A later public release may use a narrower
source-only boundary where redistribution terms require it.

\section{Artifact Inventory and Reproduction Procedure}
\label{sec:artifacts}

\begin{table*}[!t]
\centering
\small
\setlength{\tabcolsep}{4pt}
\begin{tabular}{p{.20\textwidth}p{.38\textwidth}p{.32\textwidth}}
\toprule
Stage & Entry point & Principal output \\
\midrule
Environment & \path{setup_cmi_environment.sh} & Conda environment \texttt{CMI} and package audit \\
Travel growth & \path{build_travel_plasticity_stability_growth.py} & L0--L6 paired JSONL and validation report \\
Composition & \path{build_composition_mechanism_dataset.py} & Four relationships at counts 0/2/4/8 \\
Mirrored stability & \path{run_mirrored_stability_experiment.py} & Stability retrieval, generation, and summaries \\
External construction & \path{build_*_travel_isomorphic.py} & Source-specific CMI JSONL and protocol audit \\
Retrieval & \path{run_external_travel_isomorphic_retrieval.py} & Lexical and dense outputs \\
IAAR training & \path{run_iaar_noise_training_ablation.py} & Clean-only/CMI selector outputs and paired deltas \\
Answer generation & \path{run_iaar_noise_training_answer_ablation.py} & Realized JSON answers and deterministic scores \\
Fixed-L3 answers & \path{run_external_fixed_l3_poison_answers.py} & BM25/Dense outputs and corrected poisoning scores \\
Encoder check & \path{run_embedding_backend_appendix.py} & Native-4096 composition rankings \\
Package validation & \path{validate_artifact.py} & Dataset counts, pairing, credentials, paths, result rows, and checksums \\
\bottomrule
\end{tabular}
\caption{Command-level reproduction map. Exact arguments and defaults are
recorded in the named scripts and \texttt{hyperparameters.md}.}
\label{tab:reproduction-map}
\end{table*}

The submitted reviewer artifact contains:
\begin{itemize}
\item \path{code/}: the paper-cutoff construction, retrieval, IAAR, generation,
evaluation, bootstrap, and encoder-robustness implementation;
\item \path{data/}: the exact transformed TravelPlanner, PERMA, LongMemEval,
HorizonBench, and MEME inputs plus construction and protocol audits;
\item \path{results/paper_tables/}: aggregate paper-facing CSVs and paired
bootstrap intervals;
\item \path{results/answer_generation/}, \path{results/fixed_l3_poisoning/},
and \path{results/encoder_robustness/}: realized per-sample outputs retained for
offline scoring and figure verification;
\item \path{docs/}: environment, dependencies, hyperparameters, seeds,
licenses, and third-party notices; and
\item \path{README.md}, \path{DATA_MANIFEST.md},
\path{validate_artifact.py}, and \path{FILE_MANIFEST.sha256}.
\end{itemize}

No external paper code or data repository is required to inspect the submitted
experiments. Model checkpoints and API services are not bundled because of
size and provider terms; exact identifiers and settings are recorded in
\path{docs/hyperparameters.md}. The bundled outputs permit answer scoring and
statistical checks without new API calls.

The recommended reproduction order is: unpack the archive; verify
\path{FILE_MANIFEST.sha256}; run \path{validate_artifact.py}; create the
\texttt{CMI} environment; execute retrieval on the included \path{data/}; use
the bundled outputs for deterministic scoring and base-cluster summaries; and
only then repeat model-backed retrieval or generation if desired. Hugging Face
and DeepSeek credentials are needed only for fresh model calls, not for the
offline audit.

The original experiments ran in the server environment \texttt{reasoning};
clean reproduction was validated in a newly created environment named
\texttt{CMI}. The captured repository HEAD was
\texttt{6cd9de14b71915e39ac742a20\allowbreak dc33785e14b6aab}, but the working tree contained
experiment changes. The submitted file manifest, rather than the Git hash
alone, identifies the code, data, and result snapshot.

No API keys, model weights, or embedding caches are included. The archive does
include the transformed paper inputs and realized per-sample outputs needed for
review. Known provenance limitations are retained explicitly: the standalone
PERMA and LongMemEval answer summaries were recovered from an audited
consolidated evidence record; unsupported absolute wrong-use values and the
unavailable LongMemEval stable-retention interval are not reconstructed.

\FloatBarrier

\end{document}